\documentclass[10pt,twocolumn,letterpaper]{article}

\usepackage{iccv}
\usepackage{times}
\usepackage{epsfig}
\usepackage{graphicx}
\usepackage{amsmath}
\usepackage{amssymb}
\usepackage{txfonts}
\usepackage{booktabs}
\usepackage{multirow}
\usepackage{pifont}

\usepackage[accsupp]{axessibility}

\newcommand{\xmark}{\ding{55}}
\newcommand{\gt}{ground-truth}

\usepackage[breaklinks=true,bookmarks=false,colorlinks]{hyperref}

\iccvfinalcopy 

\def\iccvPaperID{****} 

\ificcvfinal\fi

\begin{document}

\title{Sensor-Guided Optical Flow}

\author{Matteo Poggi\hspace*{1cm} Filippo Aleotti \hspace*{1cm} Stefano Mattoccia\\
Department of Computer Science and Engineering (DISI)\\
University of Bologna, Italy\\
{\tt\small \{m.poggi, filippo.aleotti2, stefano.mattoccia \}@unibo.it}
}

\twocolumn[{
\renewcommand\twocolumn[1][]{#1}
\maketitle
\newcommand{\bracketbox}[2]{%
\begin{minipage}[t]{#1}\centering%
$\underbracket[1pt][0.5mm]{\hspace{\dimexpr(#1-0.5cm)}}_{\substack{\vspace{-3.0mm}\\\colorbox{white}{~#2~}}}$%
\end{minipage}}%

\begin{center}
\includegraphics[width=0.25\textwidth]{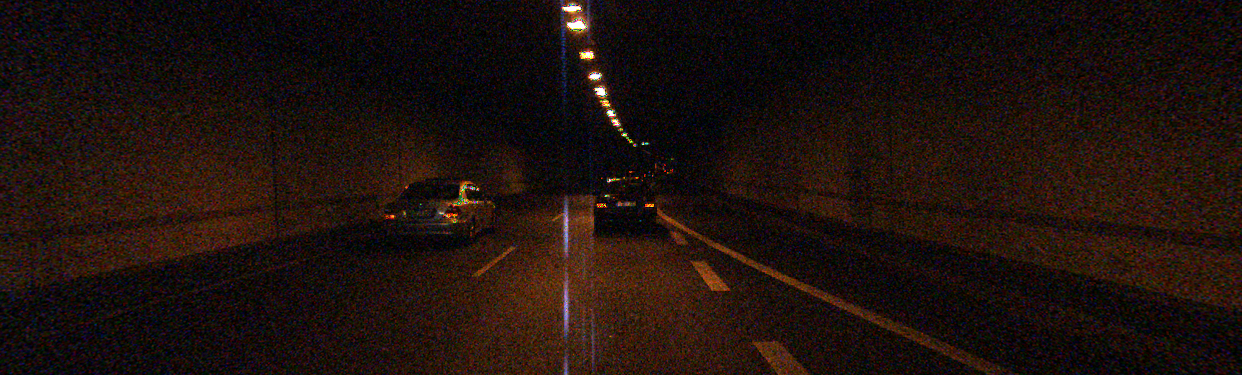} 
\includegraphics[width=0.25\textwidth]{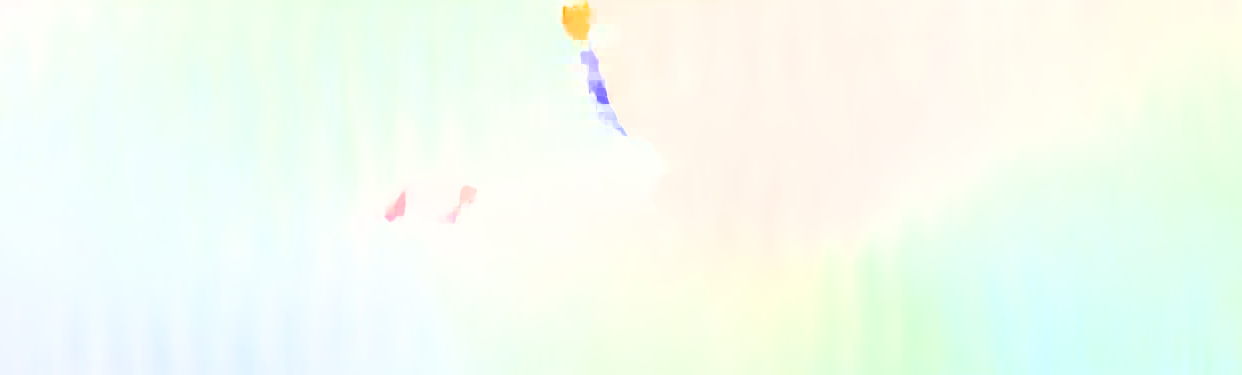} 
\includegraphics[width=0.25\textwidth]{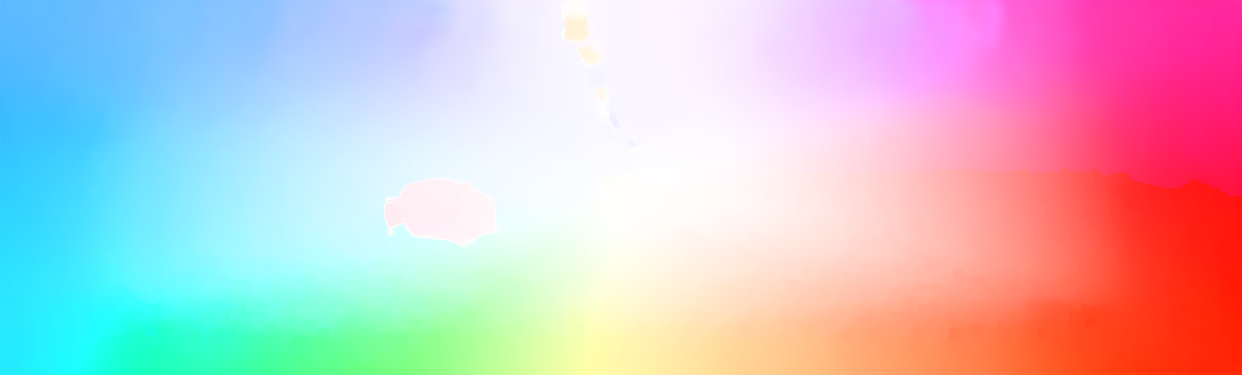} \\
\includegraphics[width=0.25\textwidth]{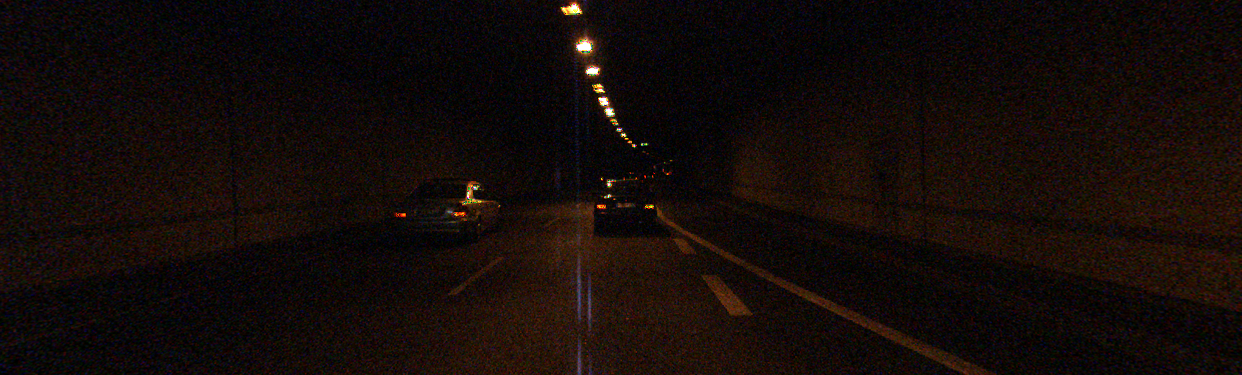} 
\includegraphics[width=0.25\textwidth]{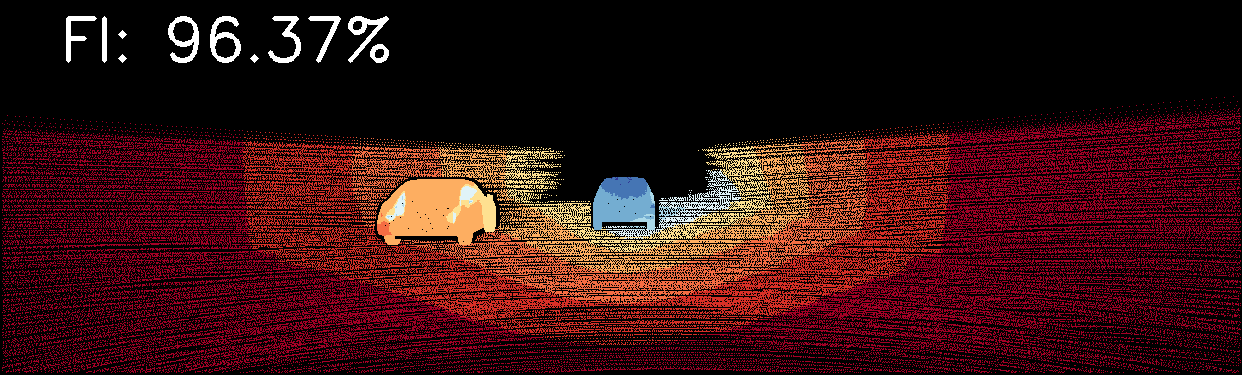} 
\includegraphics[width=0.25\textwidth]{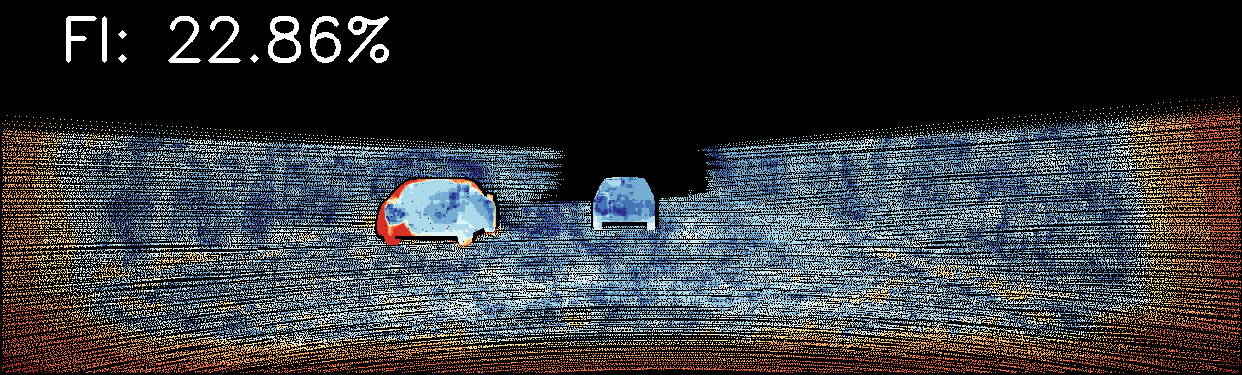} \\
\vspace{-0.1cm}
\end{center}
\vspace{-0.4cm}
\hspace{0.22\textwidth} (a) \hspace{0.23\textwidth} (b) \hspace{0.22\textwidth} (c) \hspace{0.22\textwidth} \textcolor{white}{.}

\small \hypertarget{fig:teaser}{Figure 1.} \textbf{Guided optical flow in action.} Column (a): reference images, columns (b,c): optical flow (top) and corresponding error maps (bottom). When facing challenging conditions at test time (a), an optical flow network alone (b) may struggle, while an external guide can make it more robust (c). Both networks in (b,c) have been trained on synthetic data only.
\vspace{0.4cm}
}]

\ificcvfinal\thispagestyle{empty}\fi

\begin{abstract}
This paper proposes a framework to guide an optical flow network with external cues to achieve superior accuracy either on known or unseen domains.
Given the availability of sparse yet accurate optical flow hints from an external source, these are injected to modulate the correlation scores computed by a state-of-the-art optical flow network and guide it towards more accurate predictions.
Although no real sensor can provide sparse flow hints, we show how these can be obtained by combining depth measurements from active sensors with geometry and hand-crafted optical flow algorithms, leading to accurate enough hints for our purpose.
Experimental results with a state-of-the-art flow network on standard benchmarks support the effectiveness of our framework, both in simulated and real conditions.
\end{abstract}

\section{Introduction}

The task of optical flow computation \cite{horn1981determining} aims at estimating the motion of pixels in a video sequence (\eg{}, in the most common settings, from two consecutive frames in time). As a result, several higher-level tasks can be faced from it, such as action recognition, tracking and more. Although its long history, optical flow remains far from being solved due to many challenges; the lack of texture, occlusions or the blurring effect introduced by high-speed moving objects make the problem particularly hard.

Indeed, the adoption of deep learning for dense optical flow estimation has represented a turning point during the years. The possibility of learning more robust pixels similarities \cite{bai2016exploiting,xu2017accurate} allowed, at first, to soften the issues above. Then the research trend in the field rapidly converged towards direct inference of the optical flow field in an end-to-end manner \cite{dosovitskiy2015flownet,ilg2017flownet2,sun2018pwc,sun2019models,hui18liteflownet,hui20liteflownet2,hui20liteflownet3,teed2020raft}, achieving both unrivaled accuracy and run time in comparison to previous approaches.
The availability of a large amount of training data annotated with \gt{} flow labels, in most cases obtained \textit{for free} on synthetic images \cite{butler2012sintel,dosovitskiy2015flownet,ilg2017flownet2}, ignited this spread. 
Common to most end-to-end networks is the use of a \textit{correlation layer} \cite{dosovitskiy2015flownet}, explicitly computing similarity scores between pixels in the two images in order to find matches, and thus flow.

This trend, however, introduced new challenges inherently connected to the learning process. Specifically, the use of synthetic images is rarely sufficient to achieve top performance on real data. As witnessed by many works in the field \cite{dosovitskiy2015flownet,ilg2017flownet2,sun2018pwc,sun2019models,hui18liteflownet,hui20liteflownet2,hui20liteflownet3,teed2020raft}, a network trained on synthetic images already excels on benchmarks such as Sintel \cite{butler2012sintel}, yet struggles at generalizing to real benchmarks such as KITTI \cite{geiger2012kitti,menze2015kitti}.
This phenomenon is known as \textit{domain-shift} and is usually addressed by fine-tuning on few real images with available \gt{}. Nevertheless, achieving generalization without fine-tuning still represents a desirable property when designing a neural network. 
The main cause triggering the domain-shift issue is the very different appearance of synthetic versus real images, with the former unable to faithfully model noise, lightning conditions and other effects usually found in the latter, as extensively supported by the literature \cite{hoffman2018cycada,murez2018image,ramirez2019learning,toldo2020unsupervised,Tonioni_2019_CVPR,zhang2020domain,Poggi_CVPR_2019,cai2020matching}.
However, it has been shown that a deep neural network can be \textit{guided} through external hints to reduce the domain-shift effect significantly. In particular, in the case of guided stereo matching \cite{Poggi_CVPR_2019}, a neural network can be conditioned during cost-volume computation with sparse depth measurements, obtained, for instance, employing a LIDAR sensor. This strategy dramatically increases generalization across domains, as well as specialization obtained after fine-tuning.

Inspired by these findings, in this paper we formulate the \textit{guided optical flow} framework. Supposing the availability of a sparse yet accurate set of optical flow values, we use them to modulate the correlation scores usually computed by state-of-the-art networks to guide them towards more accurate results. To this aim, we first extend the guided stereo formulation to take into account 2D cost surfaces. Then, we empirically study how the effect of the sparse points is affected by the resolution at which the correlation scores are computed and, consequently, revise the state-of-the-art flow network, RAFT \cite{teed2020raft}, to make it better leverage such a guide.
The effectiveness of this approach is evaluated, at first, from a theoretical point of view by sampling a low amount of \gt{} flow points (about 3\%) -- perturbed with increasing intensity of noise -- to guide the network, and then using flow hints obtained by a \textit{real setup}. 
However, in contrast to stereo/depth estimation \cite{Poggi_CVPR_2019}, sensors capable of measuring optical flow do not exist at all. Consequently, we show how to obtain such a sparse guide out of an active depth sensor combined with a hand-crafted flow method and an instance-segmentation network \cite{he2017maskrcnn}.
It is worth noting that the setup needed by our proposal is already regularly deployed in many practical applications, such as autonomous driving, and nowadays even available in most consumer devices like smartphones and tablets equipped with cameras and active depth sensors.

Figure \hyperlink{fig:teaser}{1} shows the potential of our method in a challenging environment (a) where the same, state-of-the-art flow network \cite{teed2020raft} has been run after being trained on synthetic images only. In its original implementation (b), the network miserably fails. Instead, the same network re-trained and guided by our framework (c) with a few hints (\eg{}, about 3\% of the total pixels, sampled from \gt{} and perturbed with random noise for this example) 
is dramatically improved. 
Experiments carried out on synthetic (FlyingChairs, FlyingThings3D, Sintel) and real (Middlebury, KITTI 2012 and 2015) datasets support our main claims:
\begin{itemize}
    \item We show, for the first time, that an optical flow network can be conditioned, or \textit{guided}, by using external cues. To this aim, we pick RAFT \cite{teed2020raft}, currently the state-of-the-art in dense optical flow estimation, and revise it to benefit from the guide at its best.
    
    \item Supposing to have the availability of less than 3\% sparse flow hints, guided optical flow allows to largely reduce the domain-shift effect between synthetic and real images, as well as to further improve accuracy on the same domain.
    
    \item Although virtually no sensor is capable of providing such accurate flow hints \cite{Menze2018JPRS}, we prove that a LIDAR sensor, combined with a hand-crafted flow algorithm, can provide a meaningful guide.

\end{itemize}

\section{Related Work}

We briefly review the literature relevant to our work.

\textbf{Hand-crafted optical flow algorithms.}
Since the seminal work by Horn and Schunck \cite{horn1981determining}, for years optical flow has been cast into an energy minimization problem \cite{brox2004high,brox2009large,brox2010large,sun2014quantitative,sun2010secrets}, for instance by means of variational frameworks \cite{black1993framework,zach2007duality}. These approaches involve a data term coupled with regularization terms, and improvements to the former \cite{brox2009large,weinzaepfel2013deepflow} or the latter \cite{ranftl2014non} have represented the primary strategy to increase optical flow accuracy for years \cite{sun2010secrets}.
While these approaches perform well in presence of small displacements, they often struggle with larger flows because of the failure of the initialization process performed by the energy minimization framework. 
Some approaches overcome this problem by interpolating a sparse set of matches \cite{leordeanu2013locally,revaud2015epicflow,li2016fast,hu2016efficient,Hu_2017_CVPR}, but they are however affected by well-known problems occurring when dealing with pixels matching, such as motion blur, violation of the brightness-consistency and so on.
More recent strategies consider optical flow as a discrete optimization problem, despite managing the sizeable 2D search space required to determine corresponding pixels between images \cite{menze2015discrete,chen2016full,xu2017accurate} is challenging.
First attempts to improve optical flow with deep networks mainly consisted of learning more robust data terms by training CNNs to match patches across images \cite{weinzaepfel2013deepflow,bai2016exploiting,xu2017accurate}, before converging to end-to-end models \cite{dosovitskiy2015flownet}.

\begin{figure*}[t]
    \centering
    \renewcommand{\tabcolsep}{15pt}
    \begin{tabular}{ccc}
        \includegraphics[width=0.19\textwidth]{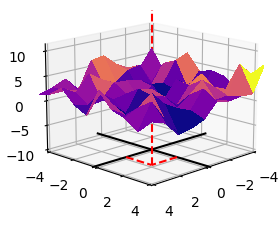}
         & 
        \includegraphics[width=0.19\textwidth]{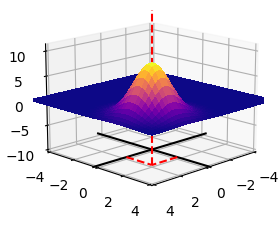} 
        &
        \includegraphics[width=0.19\textwidth]{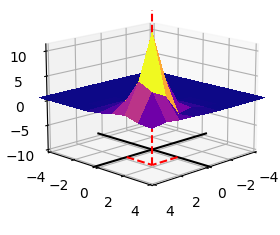} 
        \\
        (a) & (b) & (c) \\
    \end{tabular}
    \caption{\textbf{Modulation of 2D correlation scores}. Given a pixel for which a flow hint of values (2,2) is available, we show (a) raw correlation scores computed in a search window of radius 4, (b) the modulating function centered at hinted coordinates and (c) modulated scores.}
    \label{fig:guide}
\end{figure*}

\textbf{End-to-end Optical Flow.} 
The switch towards fully learnable models for estimating optical flow represented a major turning point in the field. FlowNet \cite{dosovitskiy2015flownet} is the first end-to-end deep network proposed for this purpose. In parallel, to satisfy the massive amount of training data required in this new setting, synthetic datasets with dense optical flow \gt{} labels were made available \cite{dosovitskiy2015flownet,mayer2016dispnet}. 
Starting with FlowNet, a number of architectures further improved accuracy on popular synthetic \cite{butler2012sintel,mayer2016dispnet} and real \cite{menze2015kitti, geiger2012kitti} benchmarks, designing 2D architectures \cite{ilg2017flownet2,ilg2018occlusions,zhao2020maskflownet,xiao2020learnable,teed2020raft}, refinement schemes \cite{hur2019iterative,wannenwetsch2020probabilistic} or, more recently, 4D networks as well \cite{yang2019vcn,wang2020displacement}. 
Among them, RAFT \cite{teed2020raft} currently represents the state-of-the-art. Concurrently, the use of deep networks also allowed to investigate on efficiency, leading to many compact models \cite{spynet2017,sun2018pwc,sun2019models,hui18liteflownet,hui20liteflownet2,hui20liteflownet3,yin2019hierarchical,bar2020scopeflow} capable of running in real-time at the cost of slightly lower accuracy, as well as self-supervised settings \cite{jason2016back,ren2017unsupervised,meister2018unflow,liu2019ddflow,liu2019selflow,liu2020learning,jonschkowski2020uflow}, sometimes combined with self-supervised monocular \cite{yin2018geonet,ranjan2019competitive,luo2019every,Chen_ICCV_2019,tosi2020distilled} or stereo \cite{wang2019unos,liu2020flow2stereo} depth estimation. Finally, some novel pipelines to automatically generate training data \cite{Aleotti_2021_CVPR,sun2021autoflow} have been designed. 

\textbf{Guided/conditioned deep learning.} Finally, a few works leverage the idea of conditioning deep features, either using learned \cite{huang2017arbitrary,courville2017modulating,park2019semantic} or geometry cues \cite{Poggi_CVPR_2019}. The former strategies consist of adaptive instance normalization \cite{huang2017arbitrary}, conditioned batch normalization \cite{courville2017modulating} or spatially adaptive normalization \cite{park2019semantic}, each one learning during training the modulating terms to be applied. 
In the latter case, external hints such as depth measurements by an active sensor are used to modulate geometric features, \eg{} deep matching costs in the case of stereo matching \cite{Poggi_CVPR_2019}. 

Inspired by \cite{Poggi_CVPR_2019}, in this paper, we extend such formulation to take into account 2D matching functions, as in the case of optical flow, whereas the guided stereo case is limited to a 1D modulation. Moreover, while for depth estimation tasks, the sparse hints can be easily sourced from active sensors, \eg{} LIDARs, virtually no sensor providing optical flow measurements exists \cite{Menze2018JPRS}. Thus, we also show how to obtain accurate enough cues suited for flow guidance out of an active depth sensor, this latter sometimes used to estimate 3D scene flow \cite{battrawy2019lidar,Gojcic_2021_CVPR} as well.
 
\section{Proposed framework}

In this section, we describe our framework for guided optical flow estimation. First, we recall the guided stereo matching formulation \cite{Poggi_CVPR_2019} as the background of our proposal, then we extend it to the case of optical flow. 

\subsection{Background: Guided Stereo Matching}

Given the availability of sparse yet accurate depth measurements coming, for instance, from a LIDAR sensor, a deep stereo network can be \textit{guided} to predict more accurate disparity maps by leveraging such measurements. This outcome is achieved by acting on a data structure, abstracted as a \textit{cost-volume}, where state-of-the-art networks store the probability of a pixel on the left image to match with the one on the right shifted by an offset $-d$.

Specifically, the depth hint associated with a generic pixel $p$ is converted into a disparity $d^*_p$ according to known camera parameters. Then, the cost-volume entry (\ie, cost-curve $\mathcal{C}_p$) for pixel $p$ is modulated using a Gaussian function centered on $d^*_p$, so that the single score of the cost-curve corresponding to the disparity $d = d^*_p$ is multiplied by the peak of the modulating function. Concerning the remaining scores, the farther they are from $d^*_p$ the more are dampened. This strategy yields a new cost-curve, $\mathcal{C}'_p$.
The modulation takes place only for pixels with a valid depth hint, while for the others, the original cost-curve $\mathcal{C}_p$ is kept. Thus, by defining a per-pixel binary mask $v$ in which $v_p=1$ if a depth measurement is available for pixel $p$, $v_p=0$ otherwise, the modulation can be expressed as:

\begin{equation}
\mathcal{C}'_p(d) = \left(1-v_p + v_p \cdot k \cdot e^{-\frac{(d-d^*_p)^2}{2c^2}}\right) \cdot \mathcal{C}_p(d)
\end{equation}
with $k$ and $c$ being respectively the height and width of the Gaussian. 
For stereo, $\mathcal{C}_p$ is often defined by means of a correlation layer \cite{mayer2016dispnet} or features concatenation / difference \cite{Kendall_2017_ICCV,khamis2018stereonet}. A similar practise is followed for optical flow, although the search domain is 2D rather than 1D. 

\subsection{Guided Optical Flow}

Similar to what is done by stereo networks, a common practise followed when designing an optical flow network is the explicit computation of correlation scores between features to encode the likelihood of matches. In most cases by means of 2D correlation layers \cite{dosovitskiy2015flownet} and, more recently, by concatenating features \cite{yang2019vcn,wang2020displacement}.
This leads to a 4D cost-volume structure, often reorganized to be processed by 2D convolutions for the sake of efficiency \cite{dosovitskiy2015flownet,ilg2017flownet2,teed2020raft}. In it, each entry for a generic pixel $p$ represents a 2D distribution of matching scores, corresponding to the 2D search range over which pixels are compared, as shown in Fig. \ref{fig:guide} (a). 

Accordingly, by assuming a sparse set of flow hints, consisting of 2D vectors $(x^*_p,y^*_p)$ for any pixel $p$, the correlation volume entry $C_p$ (\ie, a correlation-surface) is modulated by means of a bivariate Gaussian function centered on $(x^*_p,y^*_p)$, for which an example is shown in Fig. \ref{fig:guide} (b) having $(x^*_p,y^*_p) = (2,2)$. As a consequence, the single score of the correlation-surface corresponding to flow $(x,y)=(x^*_p,y^*_p)$ results peaked, while the remaining scores are dampened according to their distance from $(x^*_p,y^*_p)$. Again, considering a binary mask $v$ encoding pixels with a valid hint, the guided optical flow modulation can be expressed as:

\begin{equation}
\mathcal{C}'_p(x,y) = \left(1-v_p + v_p \cdot k \cdot e^{-\frac{(x-x^*_p)^2+(y-y^*_p)^2}{2c^2}}\right) \cdot \mathcal{C}_p(x,y)
\end{equation}
The resulting correlation-surface is shown in Fig. \ref{fig:guide} (c).
Although any differentiable function would be amenable for modulation, the choice of a Gaussian allows for peaking correlation scores corresponding to the hinted values together with neighboring scores, thus taking into account slight deviations of the hint from the actual flow value.

\section{Implementing Sensor-Guided Optical Flow}

As shown before, in theory, we can seamlessly extend the original stereo formulation to the optical flow problem. However, some major issues arise during the implementation. In particular, 1) existing optical flow architectures are not suited for guided optical flow and 2) obtaining flow hints from a sensor is not as natural as in the case of depth estimation, since do not exist equivalent devices capable of measuring the optical flow.
In the reminder, we will describe how to address both problems.

\subsection{Network choice and modifications}

To effectively guide the neural network to predict more accurate flow vectors, consistently with stereo formulation \cite{Poggi_CVPR_2019} we act on the similarity scores computed by specific layers of the flow networks. The literature is rich of architectures leveraging 2D correlation layers \cite{dosovitskiy2015flownet,ilg2017flownet2,sun2018pwc,hui18liteflownet,teed2020raft} or, more recently, features concatenation in 4D volumes \cite{yang2019vcn,wang2020displacement}.
Currently, RAFT \cite{teed2020raft} represents the state-of-the-art in the field and thus the preferred choice to be enhanced by our guided flow formulation, in particular, because of 1) its capacity of computing matching scores between all pairs of pixels in the two images, 2) its much faster convergence and 3) its superior generalization capability and accuracy.

However, RAFT and all the networks mentioned before usually compute correlations / concatenate features at low resolution, \ie{} $\frac{1}{8}$ or lower. On the one hand, this does not allow for a fine modulation since a single flow hint would modulate a distribution of coarse 2D correlation scores, making guided flow poorly effective or even harmful for the network, as we will see in our experiments.
On the other hand, the guided stereo framework \cite{Poggi_CVPR_2019} proved to be effective when correlation / concatenation is performed on features at $\frac{1}{4}$ resolution. Accordingly, we revise RAFT to make it suited for guided flow as follows: 1) the encoder is modified to extract features at quarter resolution, by changing the stride factor from 2 to 1 in the sixth convolutional layer and reducing the amount of extracted features from 128 to 96 to reduce complexity and memory requirements; 2) to perform convex upsampling of the predicted flow, a $\frac{\text{H}}{4} \times \frac{\text{W}}{4} \times (4\times4\times9)$ mask is predicted instead of $\frac{\text{H}}{8} \times \frac{\text{W}}{8} \times (8\times8\times9)$. We dub this Quarter resolution RAFT variant QRAFT. Experimentally, we will show that it is much better suited to leverage guided flow, significantly improving accuracy when fed with hints.

Although similar modifications are theoretically applicable to most state-of-the-art optical flow networks, they result practically unfeasible on 4D networks \cite{yang2019vcn,wang2020displacement} because of 1) the much higher complexity/memory requirements of 4D convolutions and 2) the resolution at which the volumes are built, usually $\frac{1}{16}$ or lower, that would require a much higher overhead to reach the desired quarter resolution.

\begin{figure}[t]
    \centering
    \renewcommand{\tabcolsep}{1pt}
    \begin{tabular}{ccccc}
        & & \rotatebox{90}{\tiny \quad (a) Frames} & \includegraphics[width=0.19\textwidth]{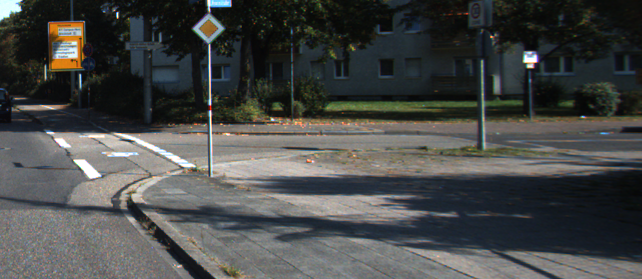}
         & 
        \includegraphics[width=0.19\textwidth]{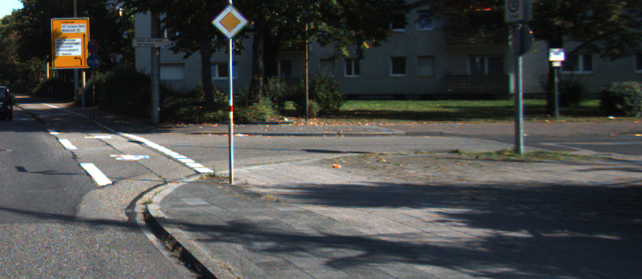} \\
        & \rotatebox{90}{\tiny \quad (b) EgoFlow --} & \rotatebox{90}{\tiny \quad no filtering} & \includegraphics[width=0.19\textwidth]{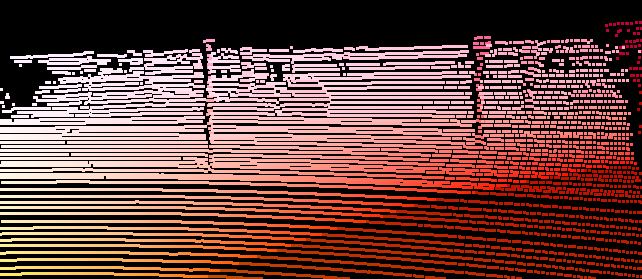}
         & 
        \includegraphics[width=0.19\textwidth]{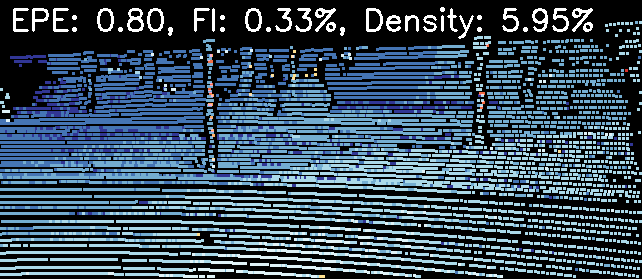} \\
        & \rotatebox{90}{\tiny \quad (c) EgoFlow --} & \rotatebox{90}{\tiny \quad filtering} & \includegraphics[width=0.19\textwidth]{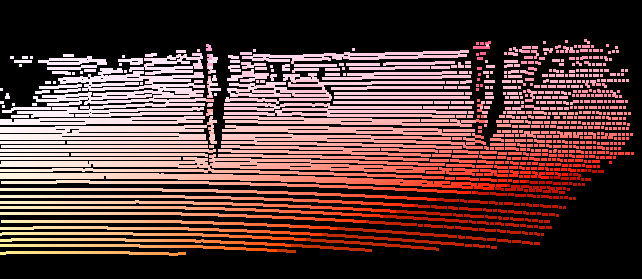}
         & 
        \includegraphics[width=0.19\textwidth]{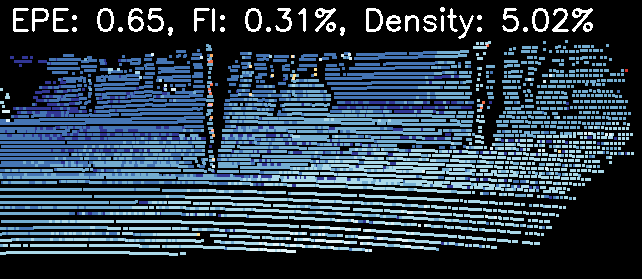} \\
    \end{tabular}
    \caption{\textbf{Optical flow hints from a LIDAR -- static scene.} Top row: reference images. The remaining rows show flow guides (left) and error maps (right), densified for better visualization. The actual density is reported over each error map, with EPE and Fl computed on pixels with both hints and ground-truth available.}
    \label{fig:lidar-static}
\end{figure}

\subsection{Accurate flow hints from active depth sensors}
\label{sec:lidarflow}

In this section, we describe a possible implementation of a real system capable of providing sparse flow guidance. Although a sensor measuring the optical flow does not exist, we can implement a virtual one by combining existing sensors and known geometry properties.
First, we point out that pixel flow between two images $\mathcal{I}_0,\mathcal{I}_1$ is the consequence of two main components: 1) camera ego-motion and 2) independently moving objects in the scene. 

\textbf{Ego-motion flow.} Concerning the former, it is straightforward to compute it by leveraging geometry if camera intrinsics $K$, depth $\mathcal{D}_0$ for pixels $p_0$ in $\mathcal{I}_0$ and relative pose $T_{0\rightarrow{}1}$ are known. Accordingly, corresponding coordinates $p_1$ in $\mathcal{I}_1$ can be obtained by projecting $p_0$ in 3D space using $K^{-1}$ and $\mathcal{D}_0$, applying roto-translation $T_{0\rightarrow{}1}$ and back-projecting to $\mathcal{I}_1$ image plane using $K$

\begin{equation}\label{eq:pix2pix}
    p_1 \sim KT_{0\rightarrow{}1}\mathcal{D}_0(p_0)K^{-1}p_0 
\end{equation}
While $K$ is known, depth $\mathcal{D}_0$ can be obtained by means of sensors, since a variety of devices for depth sensing exist, a LIDAR for instance. Finally, the relative pose $T_{0\rightarrow{}1}$ can be obtained by solving the Perspective-n-Point (PnP) problem \cite{lepetit2009epnp} between frames $\mathcal{I}_0$ and $\mathcal{I}_1$, by knowing corresponding LIDAR depths $\mathcal{D}_0$ and $\mathcal{D}_1$ and using matched feature correspondences extracted from $\mathcal{I}_0$ and $\mathcal{I}_1$, filtered by means of RANSAC \cite{fischler1981random} as in \cite{ma2018self}. Finally, flow $f^e_{0\rightarrow{}1}$ -- or \textit{EgoFlow} -- can be obtained by subtracting $p_0$ coordinates from $p_1$.

Although noisy, LIDAR measurements are accurate enough to allow for computing meaningful flow guide when dealing with static scenes, as shown in Fig. \ref{fig:lidar-static} (b). Moreover, we can further remove noisy flow estimates by deploying a forward-backward consistency mask $c^e_{0\leftrightarrow{}1}$. This is obtained by computing the ego-motion backward flow $f^e_{1\rightarrow{}0}$, by backward warping $f^e_{1\rightarrow{}0}$ according to $f^e_{0\rightarrow{}1}$ and then by comparing warped flow $\tilde{f}^e_{1\rightarrow{}0}$ with $f^e_{0\rightarrow{}1}$ itself, resulting consistent if the two flows for a same pixel $p_0$ are opposite. Thus, we consider valid pixels those having an Euclidean distance between $f^e_{0\rightarrow{}1}$ and $-\tilde{f}^e_{1\rightarrow{}0}$ lower than a threshold (\eg{}, 3):

\begin{equation}\label{eq:fwbw}
    c^e_{0\leftrightarrow{}1}(p_0) =
\begin{cases}
1 & \text{if } \quad || f^e_{0\rightarrow{}1}(p_0) + \tilde{f}^e_{1\rightarrow{}0}(p_0) ||_2 \leq 3 \\
0 & \text{otherwise } \\
\end{cases}
\end{equation}
However, since LIDAR points are sparse, they would rarely match after warping. Thus, we apply a simple completion filter based on classical image processing techniques \cite{ku2018defense} and compute $c^e_{0\leftrightarrow{}1}$ by replacing depth maps in Eq. \ref{eq:pix2pix} with their densified counterparts.
This allows to discard noisy measurements and increase the quality of the flow guide at the expense of density, as shown in Fig. \ref{fig:lidar-static} (c).
Nonetheless, this strategy alone cannot deal with dynamic objects.

\begin{figure}[t]
    \centering
    \renewcommand{\tabcolsep}{1pt}
    \begin{tabular}{ccccc}
        & & \rotatebox{90}{\tiny \quad (a) Frames} & \includegraphics[width=0.19\textwidth]{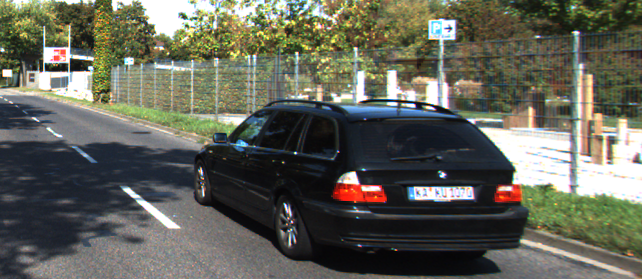}
         & 
        \includegraphics[width=0.19\textwidth]{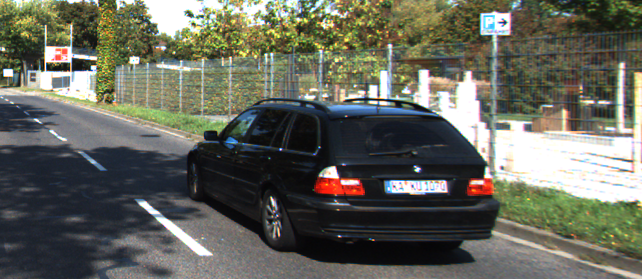} \\
        
        & \rotatebox{90}{\tiny \quad (b) EgoFlow -- } & \rotatebox{90}{\tiny \quad no filtering} & \includegraphics[width=0.19\textwidth]{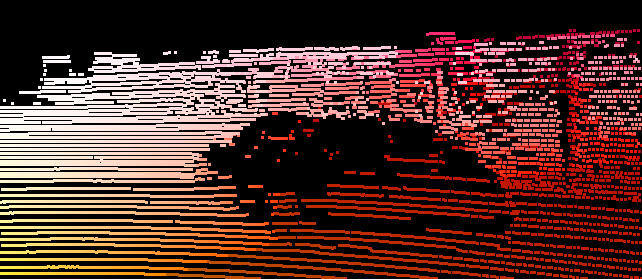} &
        \includegraphics[width=0.19\textwidth]{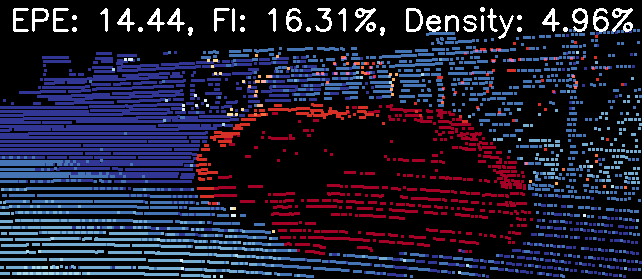} \\
        
        & \rotatebox{90}{\tiny \quad (c) EgoFlow -- } & \rotatebox{90}{\tiny \quad filtering} & \includegraphics[width=0.19\textwidth]{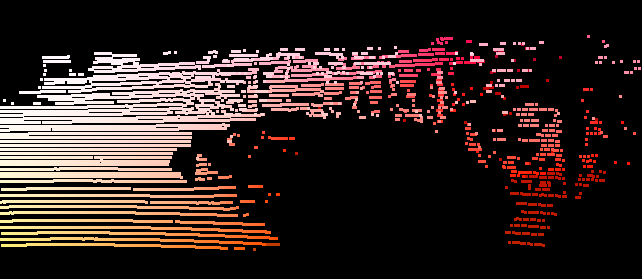} &
        \includegraphics[width=0.19\textwidth]{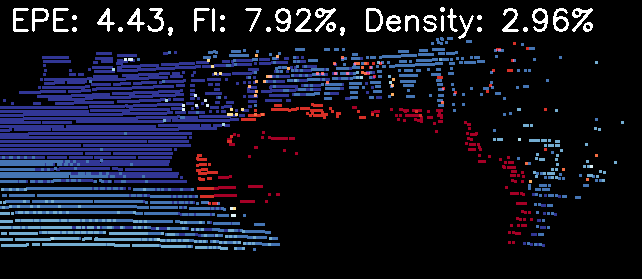} \\        
        
        & & \rotatebox{90}{\tiny \quad (d) RICFlow} & \includegraphics[width=0.19\textwidth]{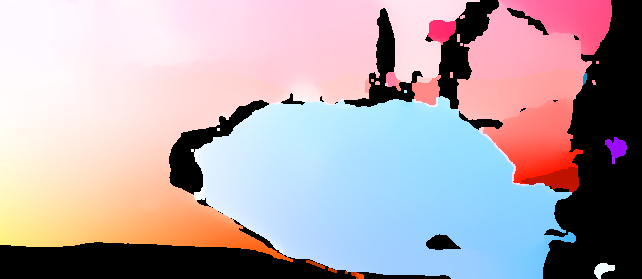}
         & 
        \includegraphics[width=0.19\textwidth]{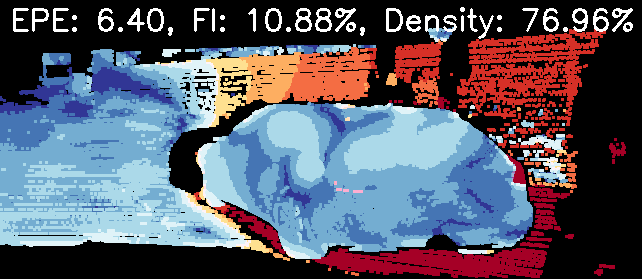} \\

        \rotatebox{90}{\tiny \quad (e) EgoFlow} & \rotatebox{90}{\tiny \quad +RICFlow} & \rotatebox{90}{\tiny \quad +MaskRCNN} & \includegraphics[width=0.19\textwidth]{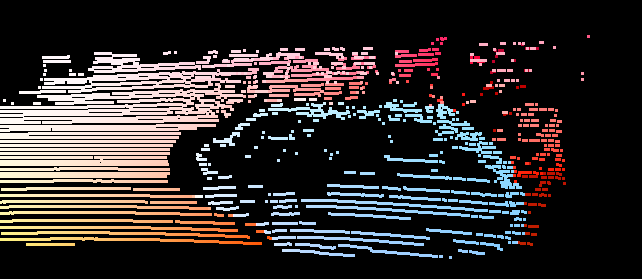} &
        \includegraphics[width=0.19\textwidth]{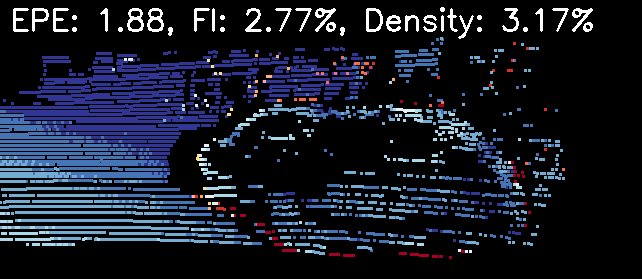} \\
    \end{tabular}
    \caption{\textbf{Optical flow hints from a LIDAR -- dynamic scene.} Top row: reference images. The remaining rows show flow guides (left) and error maps (right), densified for better visualization. The actual density is reported over each error map, with EPE and Fl computed on pixels with both hints and ground-truth available.}
    \label{fig:lidar-dynamic}
\end{figure}

\textbf{Independently moving objects flow.} The methodology introduced so far is effective when framing a static scene, but it results insufficient when moving objects appear. Indeed, Fig. \ref{fig:lidar-dynamic} shows an example in which a car is moving in the scene (a), whose flow estimated from LIDAR alone is largely incorrect, as shown in (b). 
Forward-backward consistency allows to filter out the moving car, but only partially as shown in (c). Moreover, this would not allow for recovering flow hints for dynamic objects, thus providing no cues to the neural network we wish to guide. 
To recover these missing cues, we leverage hand-crafted optical flow algorithms that indiscriminately process static and dynamic parts of the scene without the need for training (thus not suffering from domain gap issues). 
Purposely, we select RICFlow \cite{Hu_2017_CVPR} as hand-crafted algorithm because of its good trade-off between accuracy and fast inference time (a few seconds on modern CPUs), compatible with state-of-the-art networks runtime.
By running RICFlow, we obtain $f^\text{RIC}_{0\rightarrow{}1}$ and eventually perform the forward-backward consistency check detailed in Eq. \ref{eq:fwbw}. The resulting flow, shown in Fig. \ref{fig:lidar-dynamic} (d), is aware of both static and dynamic elements in the scene, although it suffers of the well-known limitations of hand-crafted algorithms, as visible for instance under the car.
However, as shown by error maps in Fig. \ref{fig:lidar-dynamic} (b) and (d), the two strategies complement each other, with LIDAR flow performing better on static regions and RICFlow on dynamic objects. Thus, we combine the two sources to obtain a complete and accurate flow guide on both cases, by distinguishing background regions from moving objects and picking EgoFlow or RICFlow accordingly. 


A strategy to achieve this task consists of explicitly segmenting the scene into background regions and foreground objects (capable of independent motion), \eg{} cars or pedestrians, for instance, employing an off-the-shelf instance segmentation network such as MaskRCNN \cite{he2017maskrcnn}. By considering the segmentation mask $s$ produced by this latter, encoding objects with different IDs, we define $f_{0\rightarrow{}1}(p_0)$ as:

\begin{equation}\label{eq:maskrcnn}
    f_{0\rightarrow{}1}(p_0) =
\begin{cases}
f^e_{0\rightarrow{}1}(p_0) & \text{if } \quad s(p_0) = 0 \\
f^\text{RIC}_{0\rightarrow{}1}(p_0) & \text{otherwise } \\
\end{cases}
\end{equation}
in which $s$ is 0 for pixels not belonging to foreground objects. This results in a guide that is meaningful on both static regions and dynamic objects, as shown in Fig. \ref{fig:lidar-dynamic} (e). 

\section{Experimental results}

In this section, we collect the outcome of our experiments. We first define the datasets involved and the implementation/training details. Then, we show: 1) a comparison between RAFT and QRAFT, 2) experiments guiding the two with sparse hints ($\sim3\%$) sampled from ground-truth or 3) with the flow guide introduced in Sec. \ref{sec:lidarflow}. 

\subsection{Datasets.}


\textbf{FlyingChairs (C) and FlyingThings3D (T).} FlyingChairs \cite{dosovitskiy2015flownet} is a popular synthetic dataset used to train optical flow models. It contains $22232$ images of chairs moving according to 2D displacement vectors over random backgrounds sampled from Flickr. The FlyingThings3D dataset \cite{ilg2017flownet2} is a collection of 3D synthetic scenes belonging to the SceneFlow dataset \cite{mayer2016dispnet} and contains a training split made of $19635$ images. Differently from C, objects move in the scene with complex 3D motions. Traditionally, both are used to pre-train flow networks: we will consider networks trained on the former only (C) or both in sequence (C+T).

\textbf{Sintel (S).} Sintel \cite{butler2012sintel} is a synthetic dataset with \gt{} optical flow maps. We use its training split, counting $1041$ images for both Clean and Final passes. In particular, we divide it into a fine-tuning split (containing sequences \textit{alley\_1, alley\_2, ambush\_2, ambush\_4, ambush\_5, ambush\_6, ambush\_7, bamboo\_1, bamboo\_2, bandage\_1, bandage\_2, cave\_2, cave\_4}) and an evaluation split (containing the remaining ones). We also evaluate networks fine-tuned on the aforementioned fine-tuning split (C+T+S).

\textbf{Middlebury Flow.} The Middlebury Flow benchmark \cite{baker2011database} is a collection of 4 synthetic and 4 real images with \gt{} optical flow maps. We use it for testing only.

\textbf{KITTI 2012 and 142 split.} The KITTI dataset is a popular dataset for autonomous driving with sparse \gt{} values for both depth and optical flow tasks. 
Two versions exist, KITTI 2012 \cite{geiger2012kitti} counting 194 images framing static scenes and KITTI 2015 \cite{menze2015kitti} made of 200 images framing moving objects, in both cases gathered by a car in motion. We use the former for evaluation only, while a split of 142 images from the latter overlaps with the KITTI raw dataset \cite{geiger2012kitti} for which raw Velodyne scans are provided, thus allowing us to validate guided flow in a real setting, namely \textit{sensor-guided optical flow}. The remaining 58 frames (K) are used in our experiments to fine-tune flow networks previously trained on synthetic data (C+T+K).

\subsection{Implementation details and training protocols.}

Our framework has been implemented starting from RAFT official source code. 
We follow the training schedules (optimizer, learning rate, iterations and weight decay) suggested in \cite{teed2020raft} to train both RAFT and QRAFT in a fair setting, training in order on C and T for 100K steps each, then fine-tuning on S or K for 50K steps. Given the higher memory requirements of QRAFT, we slightly change the crop sizes to $320\times496$, $320\times640$, $400\times720$ and $288\times960$ respectively for C, T, S and K, using image batches of 2, 1, 1 and 1, in order to fit into a single Titan Xp GPU. 
When turning on guided flow, we set $k=10$ and $c=1$ following \cite{Poggi_CVPR_2019}. The modulation acts on the correlation map computed between all pixels by downsampling the flow hints to the proper resolution with nearest neighbor interpolation.
During training, flow guide is obtained by randomly sampling 1\% pixels from the ground-truth and applying random uniform noise $\varepsilon \in [-1, 1]$, in order to make the network robust to inaccurate flow hints at test time. An ablation study on these hyper-parameters is reported in the supplementary material. 
Our demo code is available at \url{https://github.com/mattpoggi/sensor-guided-flow}.

\begin{table}[t]
\centering
\renewcommand{\tabcolsep}{5pt}
\scalebox{0.62}
{
\begin{tabular}{lll|cc|c|cc|cc}
\toprule
& Training & & \multicolumn{2}{c|}{{Sintel}} & Midd. & \multicolumn{2}{c|}{{KITTI 2012}} & \multicolumn{2}{c}{{KITTI 142}}\\
& Dataset & Network & Clean & Final & EPE & EPE & F1 & EPE & F1 \\
\midrule    
(a) & C & RAFT & 2.30 & 3.70 & 0.69 & \textbf{5.26} & 29.88 & 10.17 & 38.00 \\
(b) & C & QRAFT & \textbf{2.03} & \textbf{3.64} & \textbf{0.49} & 5.54 & \textbf{25.96} & \textbf{9.61} & \textbf{32.50} \\
\midrule
(c) & C+T & RAFT & 1.73 & 2.55 & 0.42 & 3.54 & 16.51 & 6.34 & 23.96 \\
(d) & C+T & QRAFT & \textbf{1.60} & \textbf{2.45} & \textbf{0.29} & \textbf{3.42} & \textbf{14.90} & \textbf{6.21} & \textbf{21.47} \\
\midrule
(e) & C+T+S & RAFT & 1.64 & 2.21 & 0.38  & 2.83 & 12.78 & 5.19 & 19.84 \\
(f) & C+T+S & QRAFT & \textbf{1.38} & \textbf{2.02} & \textcolor{red}{\textbf{0.27}} & \textbf{2.74} & \textbf{11.27} & \textbf{5.02} & \textbf{17.53} \\
\midrule
(g) & C+T+K & RAFT & 7.07 & 10.77 & 0.77 & \textbf{1.59} & 6.11 & 3.09 & 8.05 \\
(h) & C+T+K & QRAFT & \textbf{5.03} & \textbf{6.26} & \textbf{0.68} & 1.60 & \textbf{5.32} & \textcolor{red}{\textbf{2.58}} & \textcolor{red}{\textbf{6.61}} \\
\midrule
\midrule
(a)\textdagger & C & RAFT & 2.09 & 3.35 & 0.72 & 5.14 & 34.68  & 8.77 & 38.78 \\
(c)\textdagger & C+T & RAFT & 1.28 & 2.01 & 0.35 & 2.40 & 10.49 & 4.14 & 15.89 \\
(e)\textdagger & C+T+S & RAFT & \textcolor{red}{\textbf{1.32}} & \textcolor{red}{\textbf{1.86}} & 0.33 & 2.06 & 8.69 & 3.80 & 14.97 \\
(g)\textdagger & C+T+K & RAFT & 4.99 & 6.15 & 0.66 & \textcolor{red}{\textbf{1.47}} & \textcolor{red}{\textbf{5.15}} & 2.83 & 6.98 \\
\midrule
\midrule
(a)\textdagger\textdagger & C & RAFT \cite{teed2020raft} & 1.99 & 3.39 & 0.68 & 4.66 & 30.54 & 7.93 & 35.01 \\
(c)\textdagger\textdagger & C+T & RAFT \cite{teed2020raft} & 1.41 & 1.90 & 0.32 & 2.15 & 9.30 & 3.69 & 14.96 \\
\bottomrule
\end{tabular}
}
\caption{\textbf{Comparison between RAFT and QRAFT.} Evaluation on Sintel selection for validation (Clean and Final), Middlebury, KITTI 2012 and KITTI 142 split. 
\textdagger{} stands for $\times3$ larger batch. \textdagger\textdagger{} stands for $\times2$ GPUs ($\times6$ larger batch). \textbf{Bold}: best results on the same training setup. \textcolor{red}{\textbf{Red}}: best overall result with single GPU. }
\label{tab:raft}
\end{table}

\subsection{Comparison between RAFT and QRAFT}

We first validate the performance of QRAFT with respect to the original RAFT architecture \cite{teed2020raft}, \ie{} without using the guide. Tab. \ref{tab:raft} collects the outcome of this comparison, carried out on Sintel, Middlebury and KITTI datasets with various training configurations.
On top, we show the results achieved by training both RAFT and QRAFT with the same batch size (\ie, 2 on C, 1 on T, S and K). We can notice how QRAFT outperforms RAFT when trained in the same setting thanks to the higher resolution at which it operates, with very few exceptions -- (a) vs (b) and (g) vs (h) on KITTI 2012 EPE. However, QRAFT adds a high computational overhead compared to RAFT. Indeed, this latter can be trained with $\times3$ larger batch size on the same hardware (marked with $\dag$). In this setting, RAFT results often better than QRAFT, except on Middlebury on most cases -- (a)\textdagger{} vs (b), (c)\textdagger{} vs (d) and (e)\textdagger{} vs (f) -- and on KITTI 142 after fine-tuning -- (g)\textdagger{} vs (h).
We report, for completeness, the accuracy of models provided by the authors \cite{teed2020raft}, although trained with $\times2$ GPUs and thus not directly comparable (marked with $\dag\dag$).
Concerning efficiency, RAFT and QRAFT run respectively at 3.10 and 1.10 FPS on KITTI images (0.32 vs 0.91 sec per inference) on a Titan Xp GPU.


\begin{table*}[t]
\centering
\scalebox{0.65}
{
\begin{tabular}{lll|rr|rr|rr|rr|rr|rr|rr}
\toprule
& Training & & \multicolumn{4}{c|}{{Sintel}} & \multicolumn{2}{c|}{Middlebury Flow} & \multicolumn{4}{c|}{{KITTI 2012}} & \multicolumn{4}{c}{{KITTI 142}}\\
& Dataset & Network & \multicolumn{2}{c}{{Clean}} & \multicolumn{2}{c|}{{Final}} & \multicolumn{2}{c|}{{EPE}} & \multicolumn{2}{c}{{EPE}} & \multicolumn{2}{c|}{{Fl (\%)}} & \multicolumn{2}{c}{{EPE}} & \multicolumn{2}{c}{{Fl (\%)}} \\
\midrule
& & & \xmark & \textit{guided} & \xmark & \textit{guided} & \xmark & \textit{guided} & \xmark &  \textit{guided} & \xmark & \textit{guided} & \xmark &  \textit{guided} & \xmark & \textit{guided} \\
\midrule    
(a)\textdagger{} & C & RAFT & 2.09 & 1.70 & 3.35 & 2.54 & 0.72 & 0.63 & 5.14 & 3.51 & 34.68 & 19.26 & 8.77 & 5.39 & 38.78 & 25.73 \\
(b) & C & QRAFT & 2.03 & \textbf{1.13} & 3.64 & \textbf{1.64} & 0.49 & \textbf{0.44} & 5.54 & \textbf{2.96} & 25.96 & \textbf{15.89} & 9.61 & \textbf{4.06} & 32.50 & \textbf{19.99} \\
\midrule
(c)\textdagger{} & C+T & RAFT & 1.28 & 1.32 & 2.01 & 1.73 & 0.35 & 0.48 & 2.40 & 2.99 & 10.49 & 15.69 & 4.14 & 4.53 & 15.89 & 21.46 \\
(d) & C+T & QRAFT & 1.60 & \textbf{0.86} & 2.45 & \textbf{1.22} & 0.29 & \textbf{0.28} & 3.42 & \textbf{2.08} & 14.90 & \textbf{8.86} & 6.21 & \textbf{3.15} & 21.47 & \textbf{13.31} \\
\midrule
(e)\textdagger{} & C+T+S & RAFT & 1.32 & 1.28 & 1.86 & 1.54 & 0.33 & 0.45 & 2.06 & 2.57 & 8.69 & 12.46 & 3.80 & 4.04 & 14.97 & 18.18 \\
(f) & C+T+S & QRAFT & 1.38 & \textcolor{red}{\textbf{0.73}} & 2.02 & \textcolor{red}{\textbf{1.01}} & 0.27 & \textcolor{red}{\textbf{0.25}} & 2.74 & \textbf{1.83} & 11.27 & \textbf{7.58} & 5.02 & \textbf{2.82} & 17.53 & \textbf{11.85} \\
\midrule
(g)\textdagger{} & C+T+K & RAFT & 4.99 & 3.35 & 6.15 & 3.95 & 0.66 & 0.70 & 1.47 & 1.84 & 5.15 & 7.13 & 2.83 & 2.83 & 6.98 & 8.74 \\
(h) & C+T+K & QRAFT & 5.03 & \textbf{1.63} & 6.26 & \textbf{2.08} & 0.68 & \textbf{0.54} & 1.60 & \textcolor{red}{\textbf{1.08}} & 5.32 & \textcolor{red}{\textbf{3.19}} & 2.58 & \textcolor{red}{\textbf{1.22}} & 6.61 & \textcolor{red}{\textbf{3.78}} \\
\bottomrule
\\
\toprule
& & & \multicolumn{4}{c|}{{Sintel}} & \multicolumn{2}{c|}{Middlebury Flow} & \multicolumn{4}{c|}{{KITTI 2012}} & \multicolumn{4}{c}{{KITTI 142}}\\
& & & Clean & Final & \multicolumn{2}{r|}{Density (\%)} & \multicolumn{1}{r|}{EPE} & Density (\%) & EPE & Fl (\%) & \multicolumn{2}{r|}{Density (\%)} & EPE & Fl (\%) & \multicolumn{2}{r}{Density (\%)} \\
\midrule
(i) & & Sampled Guide & 2.30 & 2.30 & \multicolumn{2}{r|}{3.00} & \multicolumn{1}{r|}{0.77} & 2.95 & 2.29 & 18.65 & \multicolumn{2}{r|}{2.83} & 2.30 & 18.12 & \multicolumn{2}{r}{2.89} \\
\bottomrule
\end{tabular}
}
\caption{\textbf{Evaluation -- Guided Optical Flow.} Evaluation on Sintel sequences selected for validation (Clean and Final), Middlebury, KITTI 2012 and KITTI 142 split. Results without (\xmark) or with (\textit{guided}) flow guide. On the bottom, (i) statistics concerning the sampled guide.}
\label{tab:guidedflow}
\end{table*}

\subsection{Guided Optical Flow -- simulated guide}

To evaluate the guided flow framework on standard datasets, we simulate the availability of sparse flow hints ($\sim 3\%$) at test time by randomly sampling from the ground-truth flow labels. Since the availability of a \textit{perfect} guide as the one obtained by sampling from ground-truth is unrealistic, we perturb both (x,y) in the sampled guide with additive random noise $\in [-3,3]$ for Sintel and KITTI, $[-1,1]$ for Middlebury (because of the much lower magnitude of flow vectors in it).
Tab. \ref{tab:guidedflow} collects the outcome of this evaluation, carried out with both RAFT and QRAFT trained on C, C+T, C+T+S and C+T+K. For RAFT, we select the models from Tab. \ref{tab:raft} that have been trained with $\times3$ batch size (\textdagger{} entries), thus comparing the two at their best given the single Titan GPU available in our experiments. 
For both networks, we report results when computing optical flow without a guide (\xmark{} entries) or when trained and evaluated in the guided flow setting (\textit{guided} entries). Row (i) shows the error and density of the sampled guide.
In the supplementary material we report experiments at varying density and noise intensity.

\textbf{Synthetic datasets.} Results on the Sintel dataset show how both RAFT and QRAFT benefit from the guide. However, QRAFT yields much larger improvements thanks to the modulation performed on correlation scores at quarter resolution rather than at eighth resolution. The accuracy of both RAFT and QRAFT gets better and better when training on more synthetic data, respectively C, C+T and C+T+S. When fine-tuning on real data (C+T+K), the error on Sintel increases because of the domain-shift. However, guiding both RAFT and QRAFT softens this effect significantly.

\textbf{Real datasets.} When considering Middlebury and KITTI datasets, we can notice how RAFT benefits from the guide when trained on C only (a), while after being trained on T (c) and S/K (e), (g) the guide results ineffective and, in most cases, leads to lower accuracy.
On the contrary, QRAFT is always improved by the guided flow framework, consistently achieving the best results on each evaluation dataset and training configuration. In particular, we can notice how guided QRAFT achieves superior generalization compared to RAFT and QRAFT (\ie{}, when trained on C, C+T or C+T+S and evaluated on KITTI 2012 and KITTI 142), as well as it improves the results even after fine-tuning on similar domains (C+T+K).

In summary, these experiments confirm the effectiveness of the guided flow framework in a pseudo-ideal case. Nonetheless, the flow hints are 1) sampled uniformly in the image and 2) perturbed with simulated noise. Although the latter introduces the non-negligible EPE and Fl shown in Tab. \ref{tab:guidedflow} (i), it cannot appropriately model what occurs in a real case like the one we are going to investigate next.




\subsection{Sensor-Guided Optical Flow}

In this section, we evaluate the guided optical flow framework in a real setting, in which the flow hints are obtained by an actual sensors suite, as the one sketched in Sec. \ref{sec:lidarflow}. For this purpose, the KITTI 142 split is the only dataset providing both LIDAR data and ground-truth flow labels that we use for this evaluation. We point out that, since the LIDAR is not available for the training data, we train by sampling the guide from \gt{} as before. For this evaluation, we consider only QRAFT, since RAFT poorly performed when guided with sampled \gt.

\begin{table}[t]
    \centering
    \scalebox{0.7}
    {
    \begin{tabular}{l|rr|rr}
        \toprule
        & \multicolumn{4}{c}{{KITTI 142}}\\
        Guide Source & EPE & Fl (\%) & \multicolumn{2}{c}{Density (\%)} \\
        \midrule
        EgoFlow -- no filtering & 3.25 & 9.72 & \multicolumn{2}{r}{3.99} \\
        EgoFlow -- filtering & 2.39 & 6.41 & \multicolumn{2}{r}{3.24} \\
        RICFlow & 2.32 & 8.68 & \multicolumn{2}{r}{85.48} \\
        EgoFlow +RICFlow +Motion Mask \cite{ranjan2019competitive} & 1.32 & 5.04 & \multicolumn{2}{r}{3.14} \\
        EgoFlow +RICFlow +Motion Prob. \cite{tosi2020distilled} & 1.22 & 4.35 & \multicolumn{2}{r}{3.09} \\
        EgoFlow +RICFlow +MaskRCNN \cite{he2017maskrcnn} & \textbf{0.80} & \textbf{2.35} & \multicolumn{2}{r}{3.16} \\
        \bottomrule
        \end{tabular}
    }
    \caption{\textbf{Flow guide accuracy.} Evaluation on KITTI 142 split for flow hints generated by using different cues.} 
    \label{tab:lidar-guide}
\end{table}

\textbf{Flow guide accuracy.} First, we quantitatively evaluate the accuracy of the flow hints produced by the techniques introduced before. Tab. \ref{tab:lidar-guide} reports the results achieved by the different approaches shown qualitatively in Fig. \ref{fig:lidar-dynamic}. Not surprisingly, the LIDAR alone (EgoFlow) produces a high number of outliers and, in general, a large EPE. As described before, properly handling dynamic objects allows us to obtain a much more reliable flow guide, as shown in the last entry of the table, used for the following evaluation. We also show that using motion masks \cite{ranjan2019competitive} or probabilities \cite{tosi2020distilled} in place of semantics \cite{he2017maskrcnn} results less effective.

Compared to guide from Tab. \ref{tab:guidedflow} (i), the LIDAR hints have lower EPE/Fl and might expect to be even more effective. However, this is not the case because of their less regular occurrence in the image, compared to the uniform distribution obtained by sampling from \gt{} and used during training (since the LIDAR guide is not available for the training data), as shown in the supplementary material. Moreover, it can also be ascribed to the different perturbations found in actual flow hints.

\begin{table}[t]
\centering
\scalebox{0.7}
{
\begin{tabular}{lll|rr|rr}
\toprule
& Training & & \multicolumn{4}{c}{{KITTI 142}}\\
& Dataset & Network & \multicolumn{2}{c}{{EPE}} & \multicolumn{2}{c}{{Fl (\%)}} \\
\midrule
& & & & \textit{sensor-} & & \textit{sensor-} \\
& & & \xmark & \textit{guided} & \xmark & \textit{guided} \\
\midrule    
(a) & C & QRAFT & 9.61 & \textbf{5.88} & 32.50 & \textbf{25.40} \\
\midrule
(b) & C+T & QRAFT & 6.21 & \textbf{4.55} & 21.47 & \textbf{17.09} \\
\midrule
(c) & C+T+S & QRAFT & 5.02 & \textbf{4.32} & 17.53 & \textbf{15.59} \\
\midrule
(d) & C+T+K & QRAFT & 2.58 & \textcolor{red}{\textbf{2.08}} & 6.61 & \textcolor{red}{\textbf{5.97}} \\
\bottomrule
\end{tabular}
}
\caption{\textbf{Evaluation of Sensor-Guided Optical Flow.} Evaluation on KITTI 142 split, without (\xmark) or with (\textit{sensor-guided}) hints.}
\label{tab:sensorguidedflow}
\end{table}

\begin{figure}[t]
    \centering
    \renewcommand{\tabcolsep}{1pt}
    \begin{tabular}{ccccc}
        & & \rotatebox{90}{\tiny \quad (a) Frames} & \includegraphics[width=0.19\textwidth]{images/labels/000018_10.png}
         & 
        \includegraphics[width=0.19\textwidth]{images/labels/000018_11.png} \\
        & \rotatebox{90}{\tiny \quad (b) C+T+S} & \rotatebox{90}{\tiny \quad\quad\quad \xmark} & \includegraphics[width=0.19\textwidth]{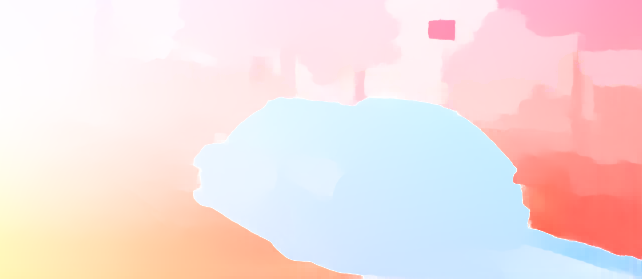}
         & 
        \includegraphics[width=0.19\textwidth]{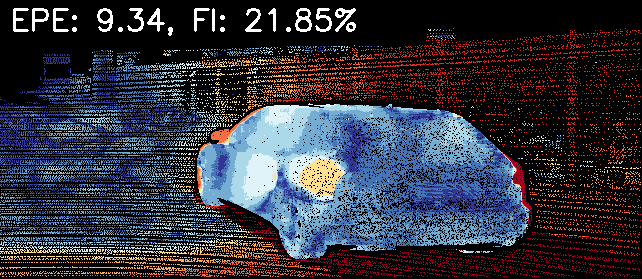} \\
        & \rotatebox{90}{\tiny \quad (c) C+T+S} & \rotatebox{90}{\tiny \quad sensor-guided} & \includegraphics[width=0.19\textwidth]{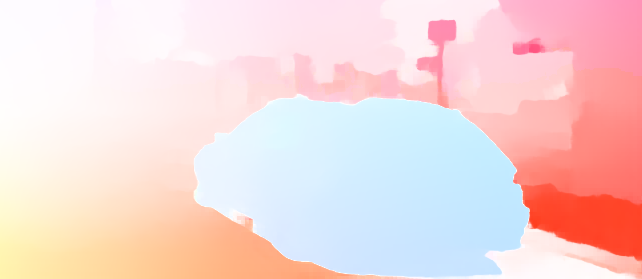}
         & 
        \includegraphics[width=0.19\textwidth]{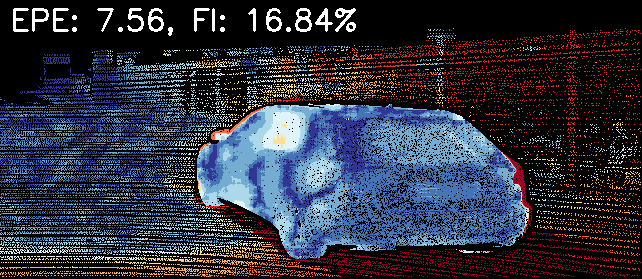} \\
    \end{tabular}
    \caption{\textbf{Qualitative results, KITTI 142 split.} From top: reference images (a) followed by flow (left) and error (right) maps by QRAFT, trained on C+T+S without (b) or with (c) sensor-guide.}
    \label{fig:raftq}
\end{figure}

\textbf{Sensor-guided QRAFT.} Once computed reliable hints, we evaluate the performance of QRAFT when guided accordingly. Tab. \ref{tab:sensorguidedflow} collects the accuracy achieved by training QRAFT in the different configurations studied so far, without (\xmark) or with guide sampled from \gt{} during training (\textit{sensor-guided}) and with the best guide selected from Tab. \ref{tab:lidar-guide} for testing. Although, for the reasons outlined before, the gain is lower compared to the use of pseudo-ideal hints 
(see Tab. \ref{tab:guidedflow} for comparison), guided QRAFT consistently beats QRAFT in any configuration.
Fig. \ref{fig:raftq} shows results by QRAFT (b) and its sensor-guided counterpart (c) both trained on C+T+S, highlighting how the guide obtained by a real system -- the one in Fig. \ref{fig:lidar-dynamic} (e) -- softens the effect due to domain-shift.

\begin{figure}[t]
    \centering
    \renewcommand{\tabcolsep}{1pt}
    \begin{tabular}{cccccccc}
        & & & \tiny RAFT \textdagger{} & \tiny RAFT \textdagger\textdagger{} \cite{teed2020raft} & \tiny QRAFT & \quad\quad & \tiny QRAFT \\
        & & & \tiny \xmark& \tiny \xmark & \tiny \xmark & \quad\quad & 
        \tiny \textit{sensor-guided} \\
        \includegraphics[width=0.06\textwidth]{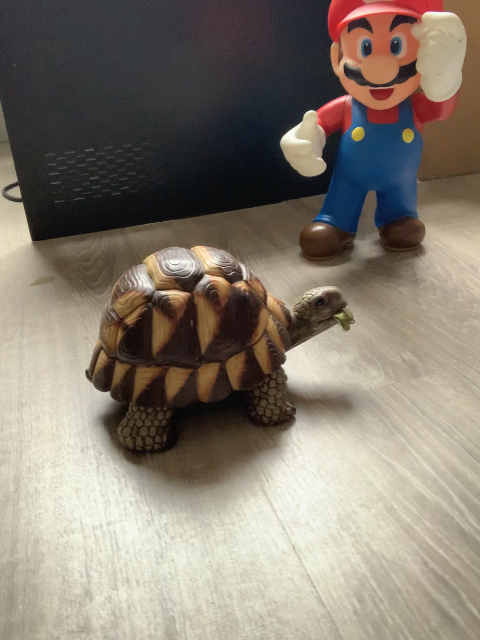} & \quad\quad &
        \rotatebox{90}{\tiny \quad\quad\quad C} &
        \includegraphics[width=0.06\textwidth]{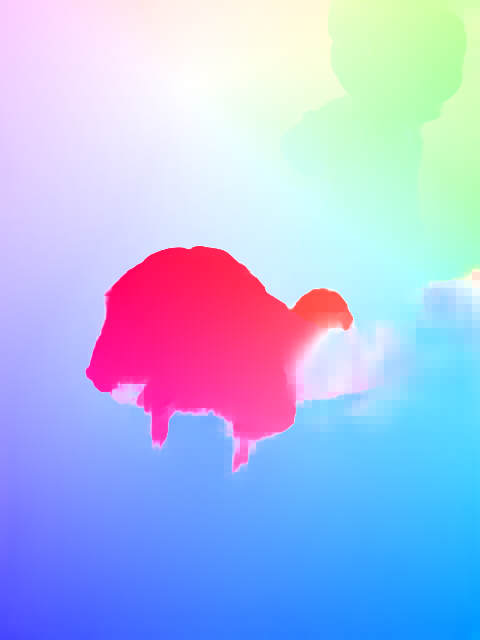} &
        \includegraphics[width=0.06\textwidth]{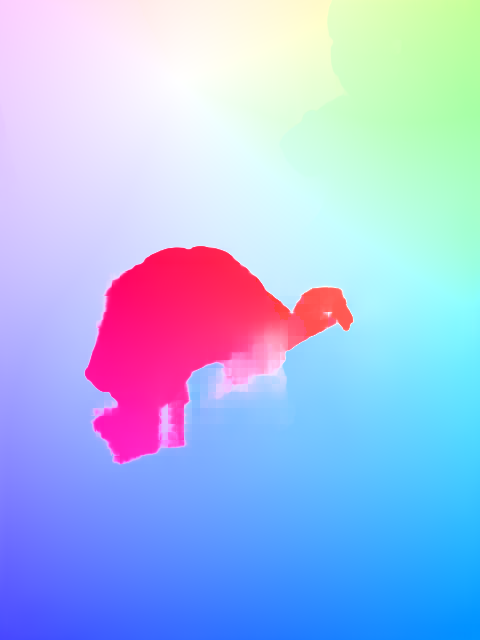} &
        \includegraphics[width=0.06\textwidth]{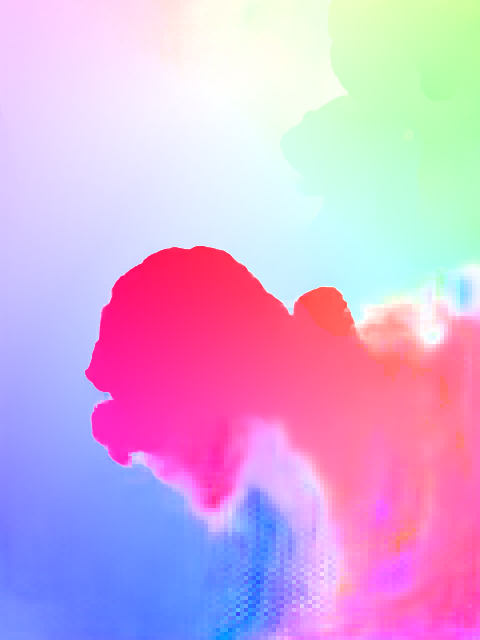} & \quad\quad &
        \includegraphics[width=0.06\textwidth]{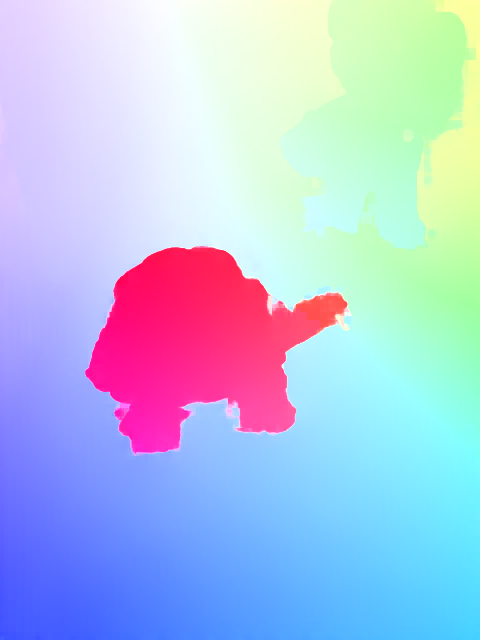} \\
        \includegraphics[width=0.06\textwidth]{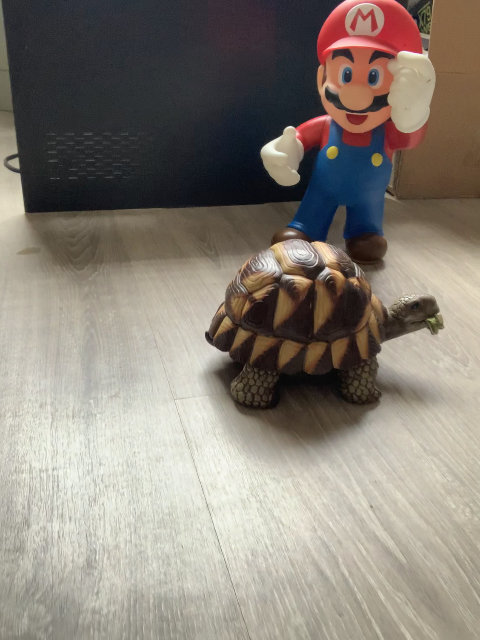} & \quad\quad &
        \rotatebox{90}{\tiny \quad\quad\quad C+T} &
        \includegraphics[width=0.06\textwidth]{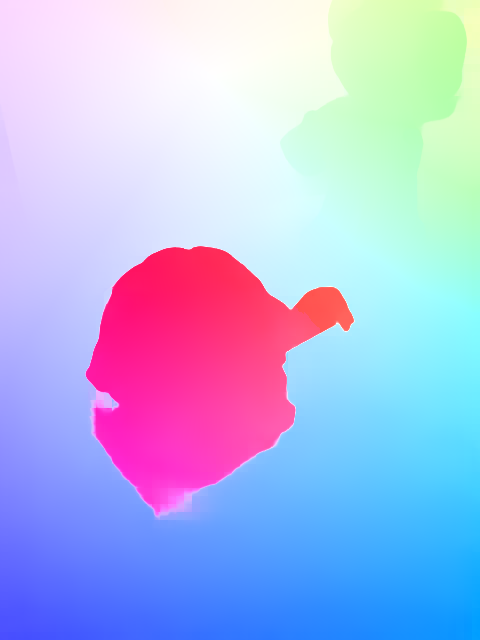} &
        \includegraphics[width=0.06\textwidth]{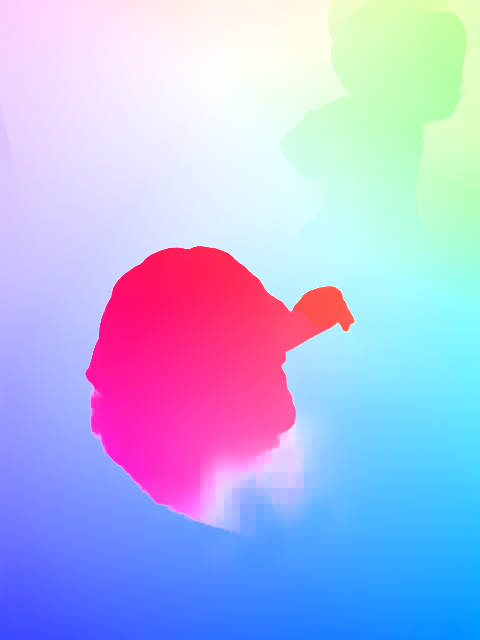} &
        \includegraphics[width=0.06\textwidth]{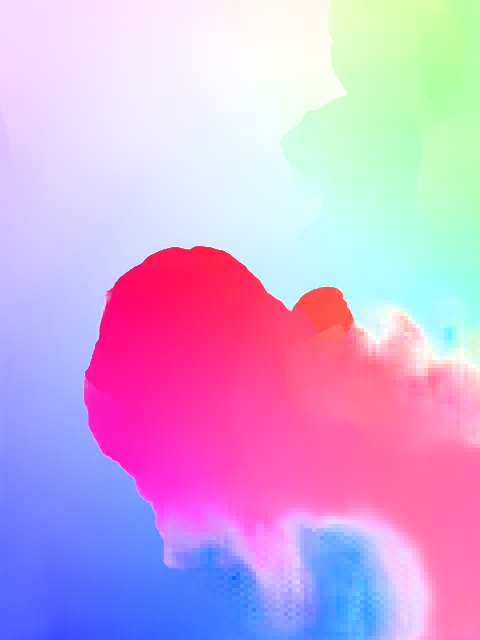} & \quad\quad &
        \includegraphics[width=0.06\textwidth]{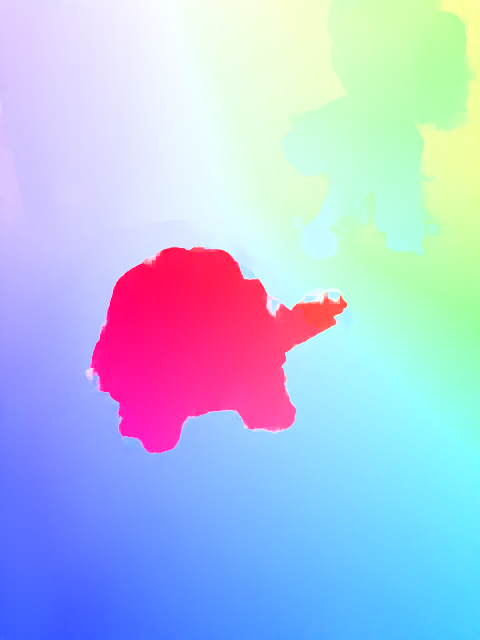} \\
    \end{tabular}
    \caption{\textbf{Qualitative results -- iPhone Xs.} On left: first (top) and second frame (bottom). On remaining columns, results by RAFT ($\times3$ batch), RAFT with authors' weights \cite{teed2020raft} ($2\times$ GPUs), QRAFT and sensor-guided QRAFT, trained on C (top) or C+T (bottom).}
    \label{fig:iphone}
\end{figure}

\textbf{Qualitative results -- handheld ToF camera.} The Velodyne used in KITTI is one among many sensors suited for sensor-guided optical flow. We show qualitatively additional results obtained with the low-res ToF sensor found in the Apple iPhone Xs, in Fig. \ref{fig:iphone}. Although on these images, QRAFT suffers more from light and shadows than RAFT, sensor-guided QRAFT vastly outperforms both. 
We report additional examples in the supplementary material.

\textbf{Limitations.} Our sensor-guided flow hints strategy is effective yet affected by some limitations. Specifically, it relies on accurate pose estimation and objects segmentation, the former performed starting from matches on images -- and thus possibly failing in the absence of distinctive features (\eg{}, large untextured regions) --  and the latter by an instance segmentation network -- failing in the presence of unknown objects. 
The failure of at least one step produces unreliable flow hints as reported in the supplementary material. 
Despite these limitations, the outcome reported in Tab. \ref{tab:sensorguidedflow} highlights clearly that sensor-guided optical flow is advantageous when a depth sensor is available, as always more often occurs in practical applications nowadays.


\section{Conclusion}

This paper has proposed a new framework, sensor-guided optical flow, that leverages flow hints to achieve better accuracy from a deep flow network. Purposely, we have revised the state-of-the-art architecture RAFT \cite{teed2020raft} to achieve superior accuracy taking advantage of our framework. We have also shown how, although a sensor measuring flow virtually does not exist \cite{Menze2018JPRS}, reliable enough flow hints can be obtained using an active depth sensor and a hand-crafted flow algorithm. Experimental results in simulated and real settings highlight the effectiveness of our proposal. 
With future advances in sensing technologies, the proposed sensor-guided optical flow can push forward further the state-of-the-art in dense flow estimation.

\textbf{Acknowledgement.} We gratefully acknowledge the support of NVIDIA Corporation with the donation of the Titan Xp GPU used for this research.

{\small
\bibliographystyle{ieee_fullname}
\bibliography{egbib}

\begin{thebibliography}{10}\itemsep=-1pt

\bibitem{Aleotti_2021_CVPR}
Filippo Aleotti, Matteo Poggi, and Stefano Mattoccia.
\newblock Learning optical flow from still images.
\newblock In {\em Proceedings of the IEEE/CVF Conference on Computer Vision and
  Pattern Recognition (CVPR)}, pages 15201--15211, June 2021.

\bibitem{bai2016exploiting}
Min Bai, Wenjie Luo, Kaustav Kundu, and Raquel Urtasun.
\newblock Exploiting semantic information and deep matching for optical flow.
\newblock In {\em European Conference on Computer Vision}, pages 154--170.
  Springer, 2016.

\bibitem{baker2011database}
Simon Baker, Daniel Scharstein, JP Lewis, Stefan Roth, Michael~J Black, and
  Richard Szeliski.
\newblock A database and evaluation methodology for optical flow.
\newblock {\em International journal of computer vision}, 92(1):1--31, 2011.

\bibitem{bar2020scopeflow}
Aviram Bar-Haim and Lior Wolf.
\newblock Scopeflow: Dynamic scene scoping for optical flow.
\newblock In {\em Proceedings of the IEEE/CVF Conference on Computer Vision and
  Pattern Recognition}, pages 7998--8007, 2020.

\bibitem{battrawy2019lidar}
Ramy Battrawy, Ren{\'e} Schuster, Oliver Wasenm{\"u}ller, Qing Rao, and Didier
  Stricker.
\newblock Lidar-flow: Dense scene flow estimation from sparse lidar and stereo
  images.
\newblock In {\em 2019 IEEE/RSJ International Conference on Intelligent Robots
  and Systems (IROS)}, pages 7762--7769. IEEE, 2019.

\bibitem{black1993framework}
Michael~J Black and Padmanabhan Anandan.
\newblock A framework for the robust estimation of optical flow.
\newblock In {\em 1993 (4th) International Conference on Computer Vision},
  pages 231--236. IEEE, 1993.

\bibitem{brox2009large}
Thomas Brox, Christoph Bregler, and Jitendra Malik.
\newblock Large displacement optical flow.
\newblock In {\em 2009 IEEE Conference on Computer Vision and Pattern
  Recognition}, pages 41--48. IEEE, 2009.

\bibitem{brox2004high}
Thomas Brox, Andr{\'e}s Bruhn, Nils Papenberg, and Joachim Weickert.
\newblock High accuracy optical flow estimation based on a theory for warping.
\newblock In {\em European conference on computer vision}, pages 25--36.
  Springer, 2004.

\bibitem{brox2010large}
Thomas Brox and Jitendra Malik.
\newblock Large displacement optical flow: descriptor matching in variational
  motion estimation.
\newblock {\em IEEE transactions on pattern analysis and machine intelligence},
  33(3):500--513, 2010.

\bibitem{butler2012sintel}
D.~J. Butler, J. Wulff, G.~B. Stanley, and M.~J. Black.
\newblock A naturalistic open source movie for optical flow evaluation.
\newblock In {A. Fitzgibbon et al. (Eds.)}, editor, {\em European Conf. on
  Computer Vision (ECCV)}, Part IV, LNCS 7577, pages 611--625. Springer-Verlag,
  Oct. 2012.

\bibitem{cai2020matching}
C. {Cai}, M. {Poggi}, S. {Mattoccia}, and P. {Mordohai}.
\newblock Matching-space stereo networks for cross-domain generalization.
\newblock In {\em 2020 International Conference on 3D Vision (3DV)}, pages
  364--373, 2020.

\bibitem{chen2016full}
Qifeng Chen and Vladlen Koltun.
\newblock Full flow: Optical flow estimation by global optimization over
  regular grids.
\newblock In {\em Proceedings of the IEEE conference on computer vision and
  pattern recognition}, pages 4706--4714, 2016.

\bibitem{Chen_ICCV_2019}
Yuhua Chen, Cordelia Schmid, and Cristian Sminchisescu.
\newblock Self-supervised learning with geometric constraints in monocular
  video: Connecting flow, depth, and camera.
\newblock In {\em ICCV}, 2019.

\bibitem{courville2017modulating}
Aaron~C Courville.
\newblock Modulating early visual processing by language.
\newblock In {\em NIPS}, 2017.

\bibitem{dosovitskiy2015flownet}
Alexey Dosovitskiy, Philipp Fischer, Eddy Ilg, Philip Hausser, Caner Hazirbas,
  Vladimir Golkov, Patrick Van Der~Smagt, Daniel Cremers, and Thomas Brox.
\newblock Flownet: Learning optical flow with convolutional networks.
\newblock In {\em Proceedings of the IEEE international conference on computer
  vision}, pages 2758--2766, 2015.

\bibitem{fischler1981random}
Martin~A Fischler and Robert~C Bolles.
\newblock Random sample consensus: a paradigm for model fitting with
  applications to image analysis and automated cartography.
\newblock {\em Communications of the ACM}, 24(6):381--395, 1981.

\bibitem{geiger2012kitti}
Andreas Geiger, Philip Lenz, and Raquel Urtasun.
\newblock Are we ready for autonomous driving? the kitti vision benchmark
  suite.
\newblock In {\em Conference on Computer Vision and Pattern Recognition
  (CVPR)}, 2012.

\bibitem{Gojcic_2021_CVPR}
Zan Gojcic, Or Litany, Andreas Wieser, Leonidas~J. Guibas, and Tolga Birdal.
\newblock Weakly supervised learning of rigid 3d scene flow.
\newblock In {\em Proceedings of the IEEE/CVF Conference on Computer Vision and
  Pattern Recognition (CVPR)}, pages 5692--5703, June 2021.

\bibitem{he2017maskrcnn}
Kaiming He, Georgia Gkioxari, Piotr Doll{\'a}r, and Ross Girshick.
\newblock Mask r-cnn.
\newblock In {\em Proceedings of the IEEE international conference on computer
  vision}, pages 2961--2969, 2017.

\bibitem{hoffman2018cycada}
Judy Hoffman, Eric Tzeng, Taesung Park, Jun-Yan Zhu, Phillip Isola, Kate
  Saenko, Alexei Efros, and Trevor Darrell.
\newblock Cycada: Cycle-consistent adversarial domain adaptation.
\newblock In {\em International conference on machine learning}, pages
  1989--1998. PMLR, 2018.

\bibitem{horn1981determining}
Berthold~KP Horn and Brian~G Schunck.
\newblock Determining optical flow.
\newblock In {\em Techniques and Applications of Image Understanding}, volume
  281, pages 319--331. International Society for Optics and Photonics, 1981.

\bibitem{Hu_2017_CVPR}
Yinlin Hu, Yunsong Li, and Rui Song.
\newblock Robust interpolation of correspondences for large displacement
  optical flow.
\newblock In {\em Proceedings of the IEEE Conference on Computer Vision and
  Pattern Recognition (CVPR)}, July 2017.

\bibitem{hu2016efficient}
Yinlin Hu, Rui Song, and Yunsong Li.
\newblock Efficient coarse-to-fine patchmatch for large displacement optical
  flow.
\newblock In {\em Proceedings of the IEEE Conference on Computer Vision and
  Pattern Recognition}, pages 5704--5712, 2016.

\bibitem{huang2017arbitrary}
Xun Huang and Serge Belongie.
\newblock Arbitrary style transfer in real-time with adaptive instance
  normalization.
\newblock In {\em Proceedings of the IEEE International Conference on Computer
  Vision}, pages 1501--1510, 2017.

\bibitem{hui20liteflownet3}
Tak-Wai Hui and Chen~Change Loy.
\newblock {LiteFlowNet3: Resolving Correspondence Ambiguity for More Accurate
  Optical Flow Estimation}.
\newblock In {\em {Proceedings of the European Conference on Computer Vision
  (ECCV)}}, 2020.

\bibitem{hui18liteflownet}
Tak-Wai Hui, Xiaoou Tang, and Chen~Change Loy.
\newblock {LiteFlowNet: A Lightweight Convolutional Neural Network for Optical
  Flow Estimation}.
\newblock In {\em {Proceedings of IEEE Conference on Computer Vision and
  Pattern Recognition (CVPR)}}, pages 8981--8989, 2018.

\bibitem{hui20liteflownet2}
Tak-Wai Hui, Xiaoou Tang, and Chen~Change Loy.
\newblock A lightweight optical flow cnn - revisiting data fidelity and
  regularization.
\newblock In {\em {IEEE Transactions on Pattern Analysis and Machine
  Intelligence}}, 2020.

\bibitem{hur2019iterative}
Junhwa Hur and Stefan Roth.
\newblock Iterative residual refinement for joint optical flow and occlusion
  estimation.
\newblock In {\em Proceedings of the IEEE/CVF Conference on Computer Vision and
  Pattern Recognition}, pages 5754--5763, 2019.

\bibitem{ilg2017flownet2}
E. Ilg, N. Mayer, T. Saikia, M. Keuper, A. Dosovitskiy, and T. Brox.
\newblock Flownet 2.0: Evolution of optical flow estimation with deep networks.
\newblock In {\em IEEE Conference on Computer Vision and Pattern Recognition
  (CVPR)}, Jul 2017.

\bibitem{ilg2018occlusions}
Eddy Ilg, Tonmoy Saikia, Margret Keuper, and Thomas Brox.
\newblock Occlusions, motion and depth boundaries with a generic network for
  disparity, optical flow or scene flow estimation.
\newblock In {\em Proceedings of the European Conference on Computer Vision
  (ECCV)}, pages 614--630, 2018.

\bibitem{jason2016back}
J~Yu Jason, Adam~W Harley, and Konstantinos~G Derpanis.
\newblock Back to basics: Unsupervised learning of optical flow via brightness
  constancy and motion smoothness.
\newblock In {\em ECCV Workshops (3)}, 2016.

\bibitem{jonschkowski2020uflow}
Rico Jonschkowski, Austin Stone, Jon Barron, Ariel Gordon, Kurt Konolige, and
  Anelia Angelova.
\newblock What matters in unsupervised optical flow.
\newblock {\em ECCV}, 2020.

\bibitem{Kendall_2017_ICCV}
Alex Kendall, Hayk Martirosyan, Saumitro Dasgupta, Peter Henry, Ryan Kennedy,
  Abraham Bachrach, and Adam Bry.
\newblock End-to-end learning of geometry and context for deep stereo
  regression.
\newblock In {\em The IEEE International Conference on Computer Vision (ICCV)},
  Oct 2017.

\bibitem{khamis2018stereonet}
Sameh Khamis, Sean Fanello, Christoph Rhemann, Adarsh Kowdle, Julien Valentin,
  and Shahram Izadi.
\newblock Stereonet: Guided hierarchical refinement for real-time edge-aware
  depth prediction.
\newblock In {\em Proceedings of the European Conference on Computer Vision
  (ECCV)}, pages 573--590, 2018.

\bibitem{ku2018defense}
Jason Ku, Ali Harakeh, and Steven~L Waslander.
\newblock In defense of classical image processing: Fast depth completion on
  the cpu.
\newblock In {\em 2018 15th Conference on Computer and Robot Vision (CRV)},
  pages 16--22. IEEE, 2018.

\bibitem{leordeanu2013locally}
Marius Leordeanu, Andrei Zanfir, and Cristian Sminchisescu.
\newblock Locally affine sparse-to-dense matching for motion and occlusion
  estimation.
\newblock In {\em Proceedings of the IEEE International Conference on Computer
  Vision}, pages 1721--1728, 2013.

\bibitem{lepetit2009epnp}
Vincent Lepetit, Francesc Moreno-Noguer, and Pascal Fua.
\newblock Epnp: An accurate o (n) solution to the pnp problem.
\newblock {\em International journal of computer vision}, 81(2):155, 2009.

\bibitem{li2016fast}
Yu Li, Dongbo Min, Minh~N Do, and Jiangbo Lu.
\newblock Fast guided global interpolation for depth and motion.
\newblock In {\em European Conference on Computer Vision}, pages 717--733.
  Springer, 2016.

\bibitem{liu2020learning}
Liang Liu, Jiangning Zhang, Ruifei He, Yong Liu, Yabiao Wang, Ying Tai, Donghao
  Luo, Chengjie Wang, Jilin Li, and Feiyue Huang.
\newblock Learning by analogy: Reliable supervision from transformations for
  unsupervised optical flow estimation.
\newblock In {\em Proceedings of the IEEE/CVF Conference on Computer Vision and
  Pattern Recognition}, pages 6489--6498, 2020.

\bibitem{liu2019ddflow}
Pengpeng Liu, Irwin King, Michael~R Lyu, and Jia Xu.
\newblock Ddflow: Learning optical flow with unlabeled data distillation.
\newblock In {\em Proceedings of the AAAI Conference on Artificial
  Intelligence}, volume~33, pages 8770--8777, 2019.

\bibitem{liu2020flow2stereo}
Pengpeng Liu, Irwin King, Michael~R Lyu, and Jia Xu.
\newblock Flow2stereo: Effective self-supervised learning of optical flow and
  stereo matching.
\newblock In {\em Proceedings of the IEEE/CVF Conference on Computer Vision and
  Pattern Recognition}, pages 6648--6657, 2020.

\bibitem{liu2019selflow}
Pengpeng Liu, Michael Lyu, Irwin King, and Jia Xu.
\newblock Selflow: Self-supervised learning of optical flow.
\newblock In {\em Proceedings of the IEEE Conference on Computer Vision and
  Pattern Recognition}, pages 4571--4580, 2019.

\bibitem{luo2019every}
Chenxu Luo, Zhenheng Yang, Peng Wang, Yang Wang, Wei Xu, Ram Nevatia, and Alan
  Yuille.
\newblock Every pixel counts++: Joint learning of geometry and motion with 3d
  holistic understanding.
\newblock {\em IEEE transactions on pattern analysis and machine intelligence},
  42(10):2624--2641, 2019.

\bibitem{ma2018self}
Fangchang Ma, Guilherme~Venturelli Cavalheiro, and Sertac Karaman.
\newblock Self-supervised sparse-to-dense: Self-supervised depth completion
  from lidar and monocular camera.
\newblock In {\em ICRA}, 2019.

\bibitem{mayer2016dispnet}
Nikolaus Mayer, Eddy Ilg, Philip Hausser, Philipp Fischer, Daniel Cremers,
  Alexey Dosovitskiy, and Thomas Brox.
\newblock A large dataset to train convolutional networks for disparity,
  optical flow, and scene flow estimation.
\newblock In {\em Proceedings of the IEEE conference on computer vision and
  pattern recognition}, pages 4040--4048, 2016.

\bibitem{meister2018unflow}
Simon Meister, Junhwa Hur, and Stefan Roth.
\newblock {UnFlow}: Unsupervised learning of optical flow with a bidirectional
  census loss.
\newblock In {\em AAAI}, 2018.

\bibitem{menze2015kitti}
Moritz Menze and Andreas Geiger.
\newblock Object scene flow for autonomous vehicles.
\newblock In {\em Conference on Computer Vision and Pattern Recognition
  (CVPR)}, 2015.

\bibitem{menze2015discrete}
Moritz Menze, Christian Heipke, and Andreas Geiger.
\newblock Discrete optimization for optical flow.
\newblock In {\em German Conference on Pattern Recognition}, pages 16--28.
  Springer, 2015.

\bibitem{Menze2018JPRS}
Moritz Menze, Christian Heipke, and Andreas Geiger.
\newblock Object scene flow.
\newblock {\em ISPRS Journal of Photogrammetry and Remote Sensing (JPRS)},
  2018.

\bibitem{murez2018image}
Zak Murez, Soheil Kolouri, David Kriegman, Ravi Ramamoorthi, and Kyungnam Kim.
\newblock Image to image translation for domain adaptation.
\newblock In {\em Proceedings of the IEEE Conference on Computer Vision and
  Pattern Recognition}, pages 4500--4509, 2018.

\bibitem{park2019semantic}
Taesung Park, Ming-Yu Liu, Ting-Chun Wang, and Jun-Yan Zhu.
\newblock Semantic image synthesis with spatially-adaptive normalization.
\newblock In {\em Proceedings of the IEEE/CVF Conference on Computer Vision and
  Pattern Recognition}, pages 2337--2346, 2019.

\bibitem{Poggi_CVPR_2019}
Matteo Poggi, Davide Pallotti, Fabio Tosi, and Stefano Mattoccia.
\newblock Guided stereo matching.
\newblock In {\em IEEE/CVF Conference on Computer Vision and Pattern
  Recognition (CVPR)}, 2019.

\bibitem{ramirez2019learning}
Pierluigi~Zama Ramirez, Alessio Tonioni, Samuele Salti, and Luigi~Di Stefano.
\newblock Learning across tasks and domains.
\newblock In {\em Proceedings of the IEEE International Conference on Computer
  Vision}, pages 8110--8119, 2019.

\bibitem{ranftl2014non}
Ren{\'e} Ranftl, Kristian Bredies, and Thomas Pock.
\newblock Non-local total generalized variation for optical flow estimation.
\newblock In {\em European Conference on Computer Vision}, pages 439--454.
  Springer, 2014.

\bibitem{spynet2017}
Anurag Ranjan and Michael~J. Black.
\newblock Optical flow estimation using a spatial pyramid network.
\newblock In {\em Proceedings of the IEEE Conference on Computer Vision and
  Pattern Recognition}, 2017.

\bibitem{ranjan2019competitive}
Anurag Ranjan, Varun Jampani, Lukas Balles, Kihwan Kim, Deqing Sun, Jonas
  Wulff, and Michael~J Black.
\newblock Competitive collaboration: Joint unsupervised learning of depth,
  camera motion, optical flow and motion segmentation.
\newblock In {\em Proceedings of the IEEE/CVF Conference on Computer Vision and
  Pattern Recognition}, pages 12240--12249, 2019.

\bibitem{ren2017unsupervised}
Zhe Ren, Junchi Yan, Bingbing Ni, Bin Liu, Xiaokang Yang, and Hongyuan Zha.
\newblock Unsupervised deep learning for optical flow estimation.
\newblock In {\em Proceedings of the AAAI Conference on Artificial
  Intelligence}, volume~31, 2017.

\bibitem{revaud2015epicflow}
Jerome Revaud, Philippe Weinzaepfel, Zaid Harchaoui, and Cordelia Schmid.
\newblock Epicflow: Edge-preserving interpolation of correspondences for
  optical flow.
\newblock In {\em Proceedings of the IEEE conference on computer vision and
  pattern recognition}, pages 1164--1172, 2015.

\bibitem{sun2010secrets}
Deqing Sun, Stefan Roth, and Michael~J Black.
\newblock Secrets of optical flow estimation and their principles.
\newblock In {\em 2010 IEEE computer society conference on computer vision and
  pattern recognition}, pages 2432--2439. IEEE, 2010.

\bibitem{sun2014quantitative}
Deqing Sun, Stefan Roth, and Michael~J Black.
\newblock A quantitative analysis of current practices in optical flow
  estimation and the principles behind them.
\newblock {\em International Journal of Computer Vision}, 106(2):115--137,
  2014.

\bibitem{sun2021autoflow}
Deqing Sun, Daniel Vlasic, Charles Herrmann, Varun Jampani, Michael Krainin,
  Huiwen Chang, Ramin Zabih, William~T Freeman, and Ce Liu.
\newblock Autoflow: Learning a better training set for optical flow.
\newblock In {\em Proceedings of the IEEE/CVF Conference on Computer Vision and
  Pattern Recognition}, pages 10093--10102, 2021.

\bibitem{sun2018pwc}
Deqing Sun, Xiaodong Yang, Ming-Yu Liu, and Jan Kautz.
\newblock Pwc-net: Cnns for optical flow using pyramid, warping, and cost
  volume.
\newblock In {\em Proceedings of the IEEE conference on computer vision and
  pattern recognition}, pages 8934--8943, 2018.

\bibitem{sun2019models}
Deqing Sun, Xiaodong Yang, Ming-Yu Liu, and Jan Kautz.
\newblock Models matter, so does training: An empirical study of cnns for
  optical flow estimation.
\newblock {\em IEEE transactions on pattern analysis and machine intelligence},
  42(6):1408--1423, 2019.

\bibitem{teed2020raft}
Zachary Teed and Jia Deng.
\newblock Raft: Recurrent all-pairs field transforms for optical flow.
\newblock In {\em European Conference on Computer Vision ({ECCV})}, 2020.

\bibitem{toldo2020unsupervised}
Marco Toldo, Andrea Maracani, Umberto Michieli, and Pietro Zanuttigh.
\newblock Unsupervised domain adaptation in semantic segmentation: a review.
\newblock {\em arXiv preprint arXiv:2005.10876}, 2020.

\bibitem{Tonioni_2019_CVPR}
Alessio Tonioni, Fabio Tosi, Matteo Poggi, Stefano Mattoccia, and Luigi~Di
  Stefano.
\newblock Real-time self-adaptive deep stereo.
\newblock In {\em Proceedings of the IEEE/CVF Conference on Computer Vision and
  Pattern Recognition (CVPR)}, June 2019.

\bibitem{tosi2020distilled}
Fabio Tosi, Filippo Aleotti, Pierluigi~Zama Ramirez, Matteo Poggi, Samuele
  Salti, Luigi Di~Stefano, and Stefano Mattoccia.
\newblock Distilled semantics for comprehensive scene understanding from
  videos.
\newblock In {\em Proceedings of the IEEE Conference on Computer Vision and
  Pattern Recognition}, 2020.

\bibitem{wang2020displacement}
Jianyuan Wang, Yiran Zhong, Yuchao Dai, Kaihao Zhang, Pan Ji, and Hongdong Li.
\newblock Displacement-invariant matching cost learning for accurate optical
  flow estimation.
\newblock {\em Advances in Neural Information Processing Systems}, 33, 2020.

\bibitem{wang2019unos}
Yang Wang, Peng Wang, Zhenheng Yang, Chenxu Luo, Yi Yang, and Wei Xu.
\newblock Unos: Unified unsupervised optical-flow and stereo-depth estimation
  by watching videos.
\newblock In {\em Proceedings of the IEEE/CVF Conference on Computer Vision and
  Pattern Recognition}, pages 8071--8081, 2019.

\bibitem{wannenwetsch2020probabilistic}
Anne~S Wannenwetsch and Stefan Roth.
\newblock Probabilistic pixel-adaptive refinement networks.
\newblock In {\em Proceedings of the IEEE/CVF Conference on Computer Vision and
  Pattern Recognition}, pages 11642--11651, 2020.

\bibitem{weinzaepfel2013deepflow}
Philippe Weinzaepfel, Jerome Revaud, Zaid Harchaoui, and Cordelia Schmid.
\newblock Deepflow: Large displacement optical flow with deep matching.
\newblock In {\em Proceedings of the IEEE international conference on computer
  vision}, pages 1385--1392, 2013.

\bibitem{xiao2020learnable}
Taihong Xiao, Jinwei Yuan, Deqing Sun, Qifei Wang, Xin-Yu Zhang, Kehan Xu, and
  Ming-Hsuan Yang.
\newblock Learnable cost volume using the cayley representation.
\newblock In {\em European Conference on Computer Vision}, pages 483--499.
  Springer, 2020.

\bibitem{xu2017accurate}
Jia Xu, Ren{\'e} Ranftl, and Vladlen Koltun.
\newblock Accurate optical flow via direct cost volume processing.
\newblock In {\em Proceedings of the IEEE Conference on Computer Vision and
  Pattern Recognition}, pages 1289--1297, 2017.

\bibitem{yang2019vcn}
Gengshan Yang and Deva Ramanan.
\newblock Volumetric correspondence networks for optical flow.
\newblock In {\em Advances in Neural Information Processing Systems 32}, pages
  794--805. Curran Associates, Inc., 2019.

\bibitem{yin2019hierarchical}
Zhichao Yin, Trevor Darrell, and Fisher Yu.
\newblock Hierarchical discrete distribution decomposition for match density
  estimation.
\newblock In {\em Proceedings of the IEEE/CVF Conference on Computer Vision and
  Pattern Recognition}, pages 6044--6053, 2019.

\bibitem{yin2018geonet}
Zhichao Yin and Jianping Shi.
\newblock Geonet: Unsupervised learning of dense depth, optical flow and camera
  pose.
\newblock In {\em Proceedings of the IEEE conference on computer vision and
  pattern recognition}, pages 1983--1992, 2018.

\bibitem{zach2007duality}
Christopher Zach, Thomas Pock, and Horst Bischof.
\newblock A duality based approach for realtime tv-l 1 optical flow.
\newblock In {\em Joint pattern recognition symposium}, pages 214--223.
  Springer, 2007.

\bibitem{zhang2020domain}
Feihu Zhang, Xiaojuan Qi, Ruigang Yang, Victor Prisacariu, Benjamin Wah, and
  Philip Torr.
\newblock Domain-invariant stereo matching networks.
\newblock In {\em European Conference on Computer Vision}, pages 420--439.
  Springer, 2020.

\bibitem{zhao2020maskflownet}
Shengyu Zhao, Yilun Sheng, Yue Dong, Eric~I Chang, Yan Xu, et~al.
\newblock Maskflownet: Asymmetric feature matching with learnable occlusion
  mask.
\newblock In {\em Proceedings of the IEEE/CVF Conference on Computer Vision and
  Pattern Recognition}, pages 6278--6287, 2020.

\end{thebibliography}
}

\end{document}